\documentclass[letterpaper]{article} 
\usepackage{aaai2026}  
\usepackage{times}  
\usepackage{helvet}  
\usepackage{courier}  
\usepackage[hyphens]{url}  
\usepackage{graphicx} 
\usepackage{natbib}  
\usepackage{caption} 
\usepackage{microtype}
\usepackage{algorithm}
\usepackage{algpseudocode}
\usepackage{bibentry}
\usepackage{bm}
\usepackage{booktabs}
\usepackage{bigstrut}
\usepackage{bbding}
\usepackage{courier}  
\usepackage{caption} 
\usepackage{enumitem}
\usepackage{float}
\usepackage{graphicx} 
\usepackage{helvet}  
\usepackage{longtable} 
\usepackage{multirow}
\usepackage{natbib}  
\usepackage{newfloat}
\usepackage{pifont}
\usepackage{subcaption}
\usepackage{times}  
\usepackage{tabularx}
\usepackage{times}
\usepackage[hyphens]{url}  
\usepackage[misc]{ifsym}
\newcommand{\xmark}{\ding{55}} 
\newcommand{\cmark}{\ding{51}}
\usepackage{amssymb}
\usepackage{newfloat}
\usepackage{listings}

 \DeclareCaptionStyle{ruled}{labelfont=normalfont,labelsep=colon,strut=off} 
\floatstyle{ruled}
\newfloat{listing}{tb}{lst}{}
\floatname{listing}{Listing}
\usepackage{nicematrix}

\makeatletter
\algdef{SE}[FOR]{ForEach}{EndForEach}[1]{%
  \algorithmicfor\ #1\ \algorithmicdo
}{%
  \algorithmicend\ \algorithmicfor
}
\makeatother

\title{MDK12-Bench: A Comprehensive Evaluation of \\ Multimodal Large Language Models on Multidisciplinary Exams}

\author{
    Pengfei Zhou\textsuperscript{1*},
    Xiaopeng Peng\textsuperscript{4*},
    Fanrui Zhang\textsuperscript{2,3*},
    Zhaopan Xu\textsuperscript{5,1},
    Jiaxin Ai\textsuperscript{6,2},
    Yansheng Qiu\textsuperscript{6,1}, \\
    Chuanhao Li\textsuperscript{1}, 
    Zhen Li\textsuperscript{1},
    Ming Li\textsuperscript{1},
    Yukang Feng\textsuperscript{2},
    Jianwen Sun\textsuperscript{2},
    Haoquan Zhang\textsuperscript{1},\\
    Zizhen Li\textsuperscript{2},
    Xiaofeng Mao\textsuperscript{1},
    Zekai Li\textsuperscript{8},
    Wangbo Zhao\textsuperscript{8},
    Kai Wang\textsuperscript{8},\\
    Xiaojun Chang\textsuperscript{3,7},
    Wenqi Shao\textsuperscript{1},
    Yang You\textsuperscript{8$\dagger$},
    Kaipeng Zhang\textsuperscript{1,2$\dagger$} \\
\hspace{-6mm} \textsuperscript{1}Shanghai AI Laboratory
    \textsuperscript{2}Shanghai Innovation Institute
    \textsuperscript{3}USTC
    \textsuperscript{4}RIT
    \textsuperscript{5}HIT
    \textsuperscript{6}WHU
    \textsuperscript{7}MBZUAI 
    \textsuperscript{8}NUS
\vspace{-0.5cm}}

\begin{document}

\maketitle

\begingroup
\renewcommand\thefootnote{}
\footnotetext{* Equal contribution. $^\dagger$ Corresponding author}
\addtocounter{footnote}{0}
\endgroup

\begin{abstract}
Multimodal large language models (MLLMs), which integrate language and visual cues for problem-solving, are crucial for advancing artificial general intelligence (AGI). However, current benchmarks for measuring the intelligence of MLLMs suffer from limited scale, narrow coverage, and unstructured knowledge, offering only static and undifferentiated evaluations. To bridge this gap, we introduce MDK12-Bench, a large-scale multidisciplinary benchmark built from real-world K–12 exams spanning six disciplines with 141K instances and 6,225 knowledge points organized in a six-layer taxonomy. Covering five question formats with difficulty and year annotations, it enables comprehensive evaluation to capture the extent to which MLLMs perform over four dimensions: 1) difficulty levels, 2) temporal (cross-year) shifts, 3) contextual shifts, and 4) knowledge-driven reasoning. We propose a novel dynamic evaluation framework that introduces unfamiliar visual, textual, and question form shifts to challenge model generalization while improving benchmark objectivity and longevity by mitigating data contamination. We further evaluate knowledge-point reference-augmented generation (KP-RAG) to examine the role of knowledge in problem-solving.  Key findings reveal limitations in current MLLMs in multiple aspects and provide guidance for enhancing model robustness, interpretability, and AI-assisted education.  
\end{abstract}

\begin{table*}[!htbp]
  \centering
  \small
   \begin{tabularx}{\textwidth}{@{}
    >{\raggedright\arraybackslash}p{3.6cm} 
    >{\centering\arraybackslash}p{1.2cm} 
    >{\centering\arraybackslash}p{1.0cm} 
    >{\centering\arraybackslash}p{3.2cm} 
    >{\centering\arraybackslash}p{1.2cm} 
    >{\centering\arraybackslash}p{1.2cm}  
    >{\centering\arraybackslash}p{0.6cm}  
    >{\centering\arraybackslash}p{0.6cm}  
    >{\centering\arraybackslash}p{0.6cm}  
    >{\centering\arraybackslash}p{0.6cm}  
    @{}}
    \toprule
    \multicolumn{1}{c}{\multirow{1}[4]{*}{Benchmarks}}  & 
    \multirow{1}[4]{*}{\#Instances} & 
    \multirow{1}[4]{*}{\#Images} & 
    \multirow{1}[4]{*}{Question Form} &
    \multirow{1}[4]{*}{\parbox{1.2cm}{\centering Answer\\Explanation}} & 
    \multirow{1}[4]{*}{\parbox{1.2cm}{\centering Knowledge\\Taxonomy}} & 
    \multirow{1}[4]{*}{\parbox{0.6cm}{\centering Diff\\Eval}} & 
    \multirow{1}[4]{*}{\parbox{0.6cm}{\centering Dyn\\Eval}}&  
    \multirow{1}[4]{*}{\parbox{0.6cm}{\centering RAG\\Eval}}&
    \multirow{1}[4]{*}{\parbox{0.6cm}{\centering Temp\\Eval}}\\\\
    \midrule
    ScienceQA \citeyearpar{lu2022learn}& 21.2K &10.3K  &  MC & \cmark &\xmark & \xmark  & \xmark& \xmark & \xmark\\
    OlympiadBench \citeyearpar{he2024olympiadbench}& 8.4K &4.8K &  Open & \xmark & \xmark& \xmark  & \xmark& \xmark & \xmark\\
    MMMU \citeyearpar{yue2024mmmu}  
          & 11.5K &12.3K &  MC, Open & \xmark & \xmark& \xmark  & \xmark& \xmark & \xmark\\
    EMMA \citeyearpar{hao2025can}  
         & 2.7K &3K& MC, Open & \xmark &\xmark& \xmark  & \xmark& \xmark& \xmark \\
    K12Vista \citeyearpar{li2025k12vista}  
         & 33K & NA &  MC, Fill, Open & \xmark &\xmark & \xmark  & \xmark& \xmark & \xmark\\
    \midrule
    MDK12-Bench 
         & \textbf{141.3K} & \textbf{105.2K}& \textbf{SC, MC, Fill, T/F, Open}& \cmark & \cmark & \cmark & \cmark& \cmark& \cmark \\
    \bottomrule
    \end{tabularx}
    \caption{\textbf{Comparison between our MDK12-Bench and existing multimodal multidisciplinary benchmarks.} MDK12-Bench includes more comprehensive question coverage, multi-layer knowledge taxonomy, and detailed explanations. It also features multi-dimensional and fine-grained evaluations including cross-difficulty evaluation (Diff Eval), dynamic evaluation (Dyn), knowledge-point reference-augmented generation evaluation (RAG) and cross-year or temporal evaluation (Temp). SC: single-choice, MC: multiple-choice; Open: Open-ended; Fill: fill-in-the-blank; T/F: true or false.}
  \label{tab:dataset_comparison}%
\end{table*}%

\section{Introduction}

Problem-solving is a core aspect of intelligence \cite{sternberg1982reasoning, lohman2011intelligence}, requiring reasoning, abstraction, knowledge integration, and adaptability to novel and increasingly difficult challenges. Achieving Artificial General Intelligence (AGI) requires models to go beyond excelling at isolated tasks, where they must demonstrate the ability to generalize knowledge, integrate information across modalities, and reason effectively under diverse contexts. Recent advances in Multimodal Large Language Models (MLLMs) ~\cite{radford2021learning, li2022blip, liu2023llava, alayrac2022flamingo} have advanced these capabilities, driving interest in rigorous benchmarks.  Measuring such capabilities demands a multidimensional and fine-grained evaluation to understand the extent to which MLLMs intelligence developed and guide improvements in model adaptability and training.





However, most existing benchmarks generally remain confined to single text modality \cite{hendrycks2measuring,hendryckstest2021,arora2023have,rein2024gpqa,zhong2024agieval} or narrow domains, such as mathematics and physics ~\citep{zhang2024mathverse,he2024olympiadbench} and medicine~\citep{sun2024pathmmu}. While efforts have been made in developing multimodal multidisciplinary benchmarks \cite{lu2022learn, yue2024mmmu, li2025k12vista, hao2025can}, they are limited in data size, diversity, and granularity. As shown in Table~\ref{tab:dataset_comparison}, most recent multidisciplinary benchmarks consists only up to 21.1K instances and limited question forms.  Additionally, the lack of fine-grained annotations, such as difficulty level and knowledge, hinders systematic evaluation of model robustness, generalization to distributional shifts, and knowledge utilization in problem solving. Moreover, static evaluations adopted by current benchmarks are susceptible to data contamination, where new benchmark data may enter the training corpora for new-generation MLLMs and become obsolete.

To address these limitations, we introduce MDK12-Bench, a large-scale, comprehensive, multidisciplinary benchmark rigorously curated from real-world K–12 exams. It is designed to assess problem-solving capabilities of MLLMs across a broad spectrum of challenges and dimensions. We systematically curated problems from six subjects: Mathematics, Physics, Chemistry, Biology, Geography, and Information Science. Following multiple rigorous filtering and processing stages, the benchmark comprises 141K unique instances collectively mapped to 6,225 human-annotated knowledge points that structured within a six-layer taxonomy. This makes our benchmark the largest of its kind. Additionally, MDK12-Bench includes five question types, instance-level difficulty annotations, and detailed answer explanations. MDK12-Bench enables fine-grained evaluation of how models progress from memorization to advanced reasoning, while capturing their generalization across task difficulty, temporal shifts, and the impact of knowledge-augmented generation.

Furthermore, we propose a dynamic evaluation framework that introduces MLLMs to unseen visual and textual shifts, providing a rigorous test of model generalization to contextual changes. This approach also promotes more objective evaluation and enhances the long-term validity of the benchmark. Additionally, we introduce knowledge-point reference-augmented generation (KP-RAG), which examines the impact of knowledge on model reasoning and problem solving.

We conducted a series of experiments on state-of-the-art MLLMs, including proprietary and open-soucrce models, evaluating both chat-oriented and reasoning-focused variants. Key findings include: \textbf{1) Model size}: Larger models outperform smaller ones across disciplines and difficulty levels, improving perception via richer multimodal representations but offering little gain in reasoning accuracy or answer completeness due to the lack of explicit reasoning-focused training and token-length limits. \textbf{2) Reasoning advantage}: Reasoning-optimized models achieve higher overall and reasoning accuracy than chat models but show limited benefits in visual perception and answer completeness. \textbf{3) Discipline difficulty}: Models struggle more in Math and Physics (7.6\% below average) than in other disciplines (3.7\% above average). \textbf{4) Harder and newer exams}: Accuracy drops by 8.3\% on harder and 12.6\% on newer exams. \textbf{5) Dynamic perturbations}: Dynamic evaluation reduces average performance by 13.7\%, revealing poor generalization. Increased sensitivity in leading models to these shifts may stem from context-aware reasoning chains or overfitting to static pretraining data. \textbf{6) Limited KP-RAG gains}: KP-RAG improves accuracy by 7\% on easy exams but only 2\% on hard ones, indicating reasoning rather than factual knowledge limits performance on harder tasks. The contributions of this work include: 
\begin{itemize}
    \item \textbf{Large-Scale, Diverse Benchmark.} We present MDK12-Bench, a multidisciplinary benchmark for evaluating MLLM problem-solving on real-world K-12 exams. It comprises 141K instances linked to 6,225 knowledge points, structured in a six-layer taxonomy, with diverse question formats, difficulty labels, and explanations.
    \item \textbf{Dynamic and Knowledge-Referenced Methods.} We propose a dynamic evaluation method to challenge generalization under contextual shifts and mitigate data contamination. We also introduce a knowledge-point reference-augmented generation (KP-RAG) pipeline to enhance answer generation and examine the role of knowledge in problem solving and reasoning.
    \item \textbf{Comprehensive Multi-Dimensional Evaluation.} We evaluate state-of-the-art MLLMs across difficulty, temporal, contextual, and knowledge dimensions, showing notable accuracy drops on harder and newer exams, as well as dynamic contextual changes. Limited KP-RAG improvement on harder tasks is also observed.
    \item \textbf{Extensive Leaderboard.} We present detailed rankings and analyses of both  proprietary and open-soucrce MLLMs. Findings on results highlight the value of the MDK-benchmark in advancing our understanding of both the capabilities and critical limitations of current MLLMs.
\end{itemize}

\section{Related Works}
\label{sec:formatting}

 \begin{figure*}[ht]
	\centering
	\includegraphics[width=0.95\textwidth]{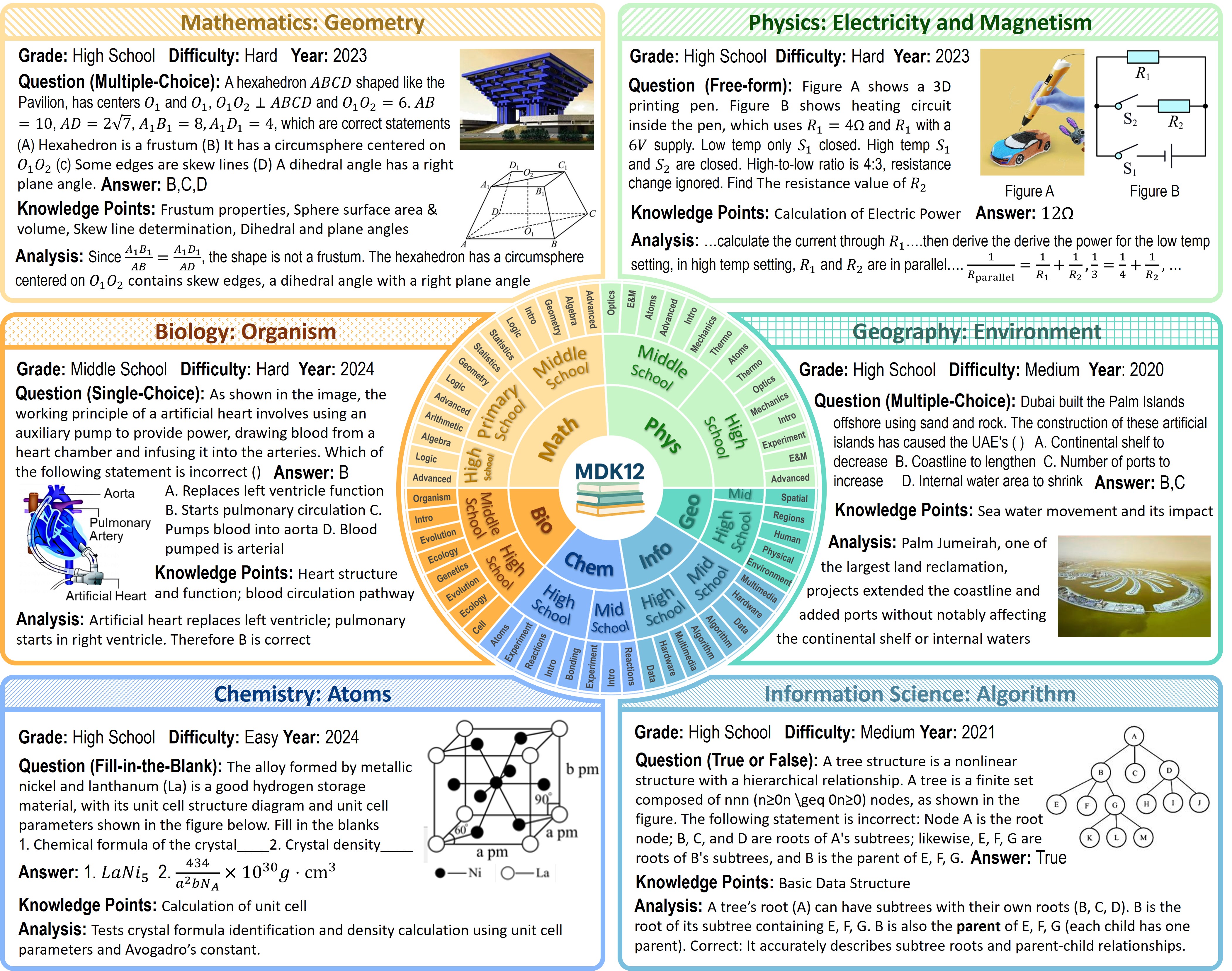}
  \caption{\textbf{Overview of MDK12-Bench.} It comprises 141K instances and spans 6 disciplines in K-12 education. Our bench is structured into a six-layer taxonomy hierarchy: subject, grade, subfield, curriculum, topics, and knowledge points, where the three rings showcase the first three layers. Examples illustrate the representative exams questions and fine-grained annotations of grade and difficulty level, exam year, question form, knowledge points and detailed analysis.}
	\label{fig:data}
\end{figure*}

\noindent\textbf{Benchmarking MLLMs.}
Evaluating the intelligence of MLLM has been challenging. The early benchmarks used text-only exams, for example, GSM-8K~\cite{cobbe2021training} and MATH~\cite{hendrycks2measuring} for math reasoning, as well as MMLU~\cite{hendryckstest2021}, JEEBench \cite{arora2023have}, GPQA \cite{rein2024gpqa} and AGIEval~\cite{zhong2024agieval} for multitask accuracy. Multimodal benchmarks later emerged, such as MathVerse~\cite{zhang2024mathverse} for visual math, SciEval \cite{sun2024scieval} for physics, MMBench~\cite{liu2024mmbench} for basic multimodal understanding, and EXAMS-V~\cite{das2024exams} for multilingual testing. More recent benchmarks, such as SciBench~\cite{wang2024scibench}, ScienceQA~\cite{lu2022learn}, MMMU~\cite{yue2024mmmu}, EMMA~\cite{hao2025can}, OlympiadBench~\cite{he2024olympiadbench}, and K12Vista~\cite{li2025k12vista}, starting to span multiple subjects. However, these efforts remain limited in scale, diversity of questions, fine-grained annotations, and multidimensional evaluation. Few addresses differentiated difficulty~\cite{ding2024easy2hard} or provide limited knowledge taxonomy~\cite{wang2024scibench, huang2024olympicarena}. Our MDK12-Bench addresses these gaps with a comprehensive large-scale dataset organized in a six-level knowledge taxonomy. It provides instance-level grade and difficulty labels, five question formats, temporal annotations, detailed knowledge points, and answer explanations.

\noindent\textbf{Dynamic Evaluation.}
Building a strong benchmark requires not only rich and well-curated data, but also carefully designed rigorous evaluation methods. As pretraining corpora expand, benchmarks face a growing risk of data contamination \cite{xu2024benchmarking, chenwe}, where test content overlaps with training data, causing leaderboards to overstate true model capabilities. Most existing benchmarks rely on static evaluations, making them particularly vulnerable to such contamination and raising doubts about their objectivity and long-term validity. To mitigate contamination, adaptive evaluation methods have been explored. Meta-probing agents~\cite{zhu2024dynamic} dynamically adjust the test content and difficulty during fine-tuning or domain adaptation, while other approaches alter visual and textual contexts to assess contamination effects~\cite{yang2024dynamic}. Beyond contamination, out-of-distribution visual changes may also affect the generalization and reasoning capabilities of MLLMs \cite{humrag}. Building on these insights, we propose a dynamic evaluation framework for MDK12-Bench that addresses both challenges. Unlike prior methods focused solely on contamination, our framework deliberately introduces novel visual, textual, and question-type variations during the test time, evaluating the generalization of models to unfamiliar conditions while ensuring sustained benchmark integrity over time.

\begin{figure*}[t]
	\centering
	\includegraphics[trim=0.5 0.5cm 0 0, width=0.98\textwidth]{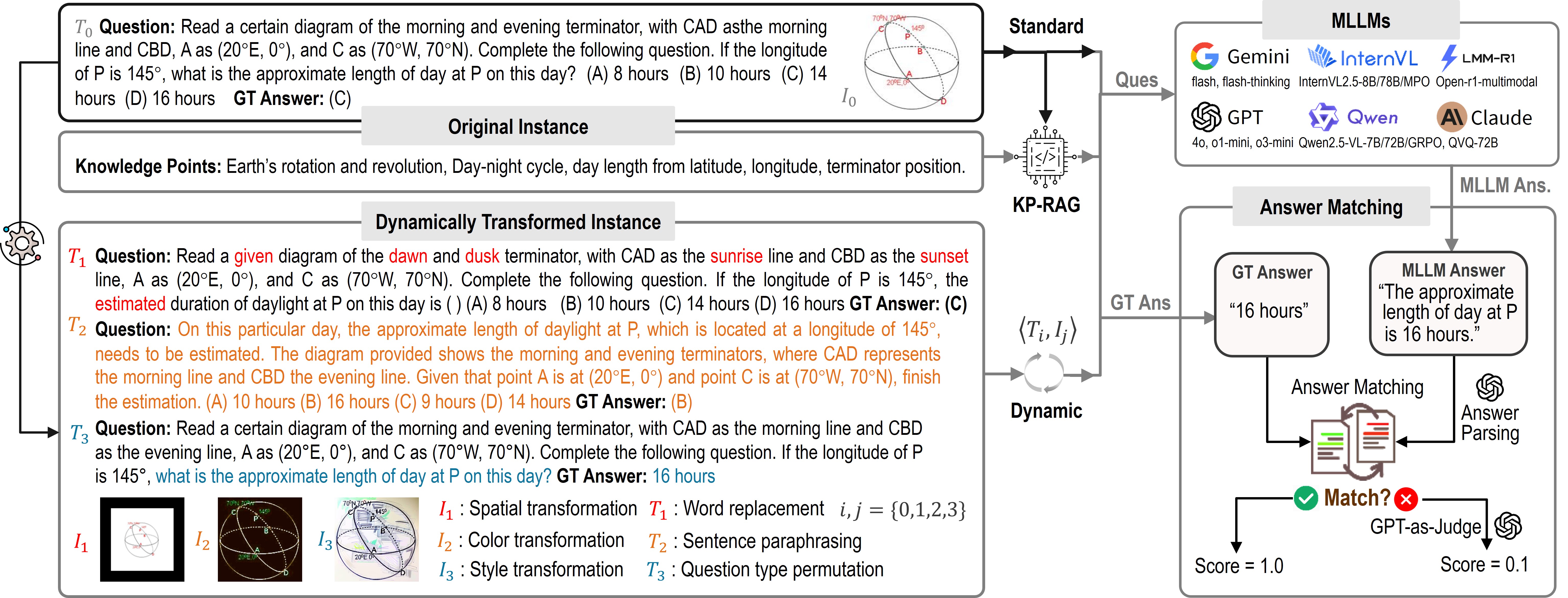}
    \caption{Our multi-dimensional evaluation pipeline comprises standard, KP-RAG, and dynamic evaluation methods.}
\label{fig:dynamic_eva}
\vspace{-0.2cm}
\end{figure*}

\begin{table}[ht]
  \centering
  \small
  \caption{Key Statistics of MDK12-Bench}
   \begin{tabularx}{0.39\textwidth}{@{}
    >{\raggedright\arraybackslash}p{3.5cm} 
    >{\centering\arraybackslash}p{3.5cm} 
    @{}}
\toprule
\multicolumn{2}{c}{\textbf{MDK-Full Statistics}} \\
\midrule
Total instances         & 141,320 \\
Text-only instances     & 77,857 (55.1\%)  \\
Multimodal instances  & 63,463 (44.9\%)  \\
Total images            & 105,218 \\
Exam years coverage & 12 years  \\
\midrule
\multicolumn{2}{c}{\textbf{MDK-Mini Statistics}} \\
\midrule
Total instances & 14,856 \\
Easy/Medium/Hard &4.952 each\\
\midrule
\multicolumn{2}{c}{\textbf{Knowledge Taxonomy Statistics}} \\
\midrule
Level 1 Disciplines  & 6 \\
Level 2 Grade & 3 (math) or 2 (other) \\
Level 3 Subfield & 36 \\
Level 4 Curriculum & 90 \\
Level 5 Topics & 499 \\
Level 6 Knowledge points & 6,225 \\
\bottomrule
\end{tabularx}
\vspace{-0.2cm}
\label{tab:dataset_stats}%
\end{table}

\section{MKD12-Benchmark}
Collecting multimodal multidisciplinary datasets is inherently challenging, which requires precise text–image pairing and domain expertise. MKD12-Bench was curated with the participation of more than 20 researchers and several K-12 educators. The process is summarized below, with illustrations and details in the Supplementary Material.


\noindent\textbf{Data Collection.}
Our data collection follows four key principles: (1) covering multidisciplinary, real-world exams; (2) incorporating diverse visual contexts and question formats for comprehensive evaluation; (3) ensuring varied difficulty levels and broad regional coverage to reveal model limitations and reduce bias; and (4) supporting robust evaluations. Guided by these principles, we gathered 5.8M multimodal exam instances from open-access K–12 repositories spanning a wide range of grades, curriculum, knowledge, question formats, and exam years.



\noindent\textbf{Data Screening.}
The data screening of the initial collection is conducted in three stages to ensure high-quality, curriculum-aligned exam data. \textbf{Rule-based Filtering}: A comprehensive set of rules, designed by data experts, is applied to automatically remove low-quality or irrelevant instances. These rules cover text–image correspondence, image resolution and clarity, content completeness, metadata accuracy, structural and formatting consistency, semantic coherence, duplication and redundancy checks, logical soundness, and appropriate year coverage. After this step, the dataset is reduced to 4.2M instances. \textbf{GPT-based Filtering}: To further refine quality, GPT-4o is employed to automatically assess semantic consistency between questions and answers, reasoning soundness, factual correctness, language clarity, and overall content completeness. This automated review filters the dataset down to 0.6M instances. \textbf{Educator Filtering}: Finally,  professional K–12 educators manually review the remaining data to ensure strict curriculum alignment, question–answer correctness, reasoning validity, and adherence to formatting standards. This validation results in a curated dataset of 0.2M instances.



\newcolumntype{Y}{>{\centering\arraybackslash}X}
\begin{table*}[ht]
  \centering
  \small
      \caption{Performance of MLLMs across six disciplines and three difficulty levels, and average over all}
  \begin{tabularx}{0.99\textwidth}{@{}
    >{\raggedright\arraybackslash}p{2.9cm} 
    >{\raggedright\arraybackslash}p{0.01cm} 
    *{18}{Y}                         
    @{ \hspace{0cm} }  
  @{}}
    \toprule
    \multicolumn{1}{l}{\multirow{3}{*}{\textbf{Models}}} 
      & \multicolumn{1}{c}{\multirow{3}{*}{\hspace{-0.25cm}\textbf{Overall}}}
      & \multicolumn{3}{c}{\textbf{Mathematics}}
      & \multicolumn{3}{c}{\textbf{Physics}}
      & \multicolumn{3}{c}{\textbf{Chemistry}}
      & \multicolumn{3}{c}{\textbf{Biology}}  
      & \multicolumn{3}{c}{\textbf{Geography}}  
      & \multicolumn{3}{c}{\textbf{Info Sci}} \\
    \cmidrule(lr){3-5}\cmidrule(lr){6-8}\cmidrule(lr){9-11}\cmidrule(lr){12-14}\cmidrule(lr){15-17}\cmidrule(lr){18-20}
& 
& \fontsize{8}{7.2}\selectfont Easy 
& \fontsize{8}{7.2}\selectfont Med 
& \fontsize{8}{7.2}\selectfont Hard 
& \fontsize{8}{7.2}\selectfont Easy 
& \fontsize{8}{7.2}\selectfont Med 
& \fontsize{8}{7.2}\selectfont Hard 
& \fontsize{8}{7.2}\selectfont Easy 
& \fontsize{8}{7.2}\selectfont Med 
& \fontsize{8}{7.2}\selectfont Hard 
& \fontsize{8}{7.2}\selectfont Easy 
& \fontsize{8}{7.2}\selectfont Med 
& \fontsize{8}{7.2}\selectfont Hard 
& \fontsize{8}{7.2}\selectfont Easy 
& \fontsize{8}{7.2}\selectfont Med 
& \fontsize{8}{7.2}\selectfont Hard 
& \fontsize{8}{7.2}\selectfont Easy 
& \fontsize{8}{7.2}\selectfont Med 
& \fontsize{8}{7.2}\selectfont Hard \\
\midrule
Gemini2-thinking & 67.8   
& 68.7&67.9&50.6   
& 69.3&61.8&54.2   
& 74.4&70.2&51.7   
& 74.1&67.4&47.2   
& 72.2&77.5&63.3   
& 81.4&77.3&73.1   
\\
GPT-o1  &  65.5
&56.2 & 50.1&58.2 
&60.9 &58.4&52.5 
&71.8&76.5&60.5 
&59.7&81.1&55.7 
&77.3&70.7&67.1 
&86.4&66.0&69.8 
\\
GPT-o1-mini & 62.4
& 53.3&47.8&55.6
& 58.1&54.3&49.2
& 68.9&71.7&57.4
& 57.4&78.2&53.4
& 73.5&67.3&64.1
& 82.3&63.4&66.5
\\
Gemini2-flash & 61.5 
& 65.8&65.3&42.1 
& 65.6&58.3&49.7 
& 69.4&65.1&43.8 
&69.0&63.1&38.9 
&67.9&73.3&58.1
&75.9&72.6&69.3
\\
GPT-4o &   59.0   
&55.1 & 53.4&45.3 
&59.3&54.7&49.6 
&64.3&64.0&60.3 
&67.2&67.8&69.2 
&66.2&57.4&53.3 
&68.4&62.6&70.1 
\\
Claude-3.7 & 58.2
& 54.4&51.2&43.3
& 56.3&53.1&47.6
& 62.4&61.3&57.7
& 64.5&65.2&66.8
& 63.6&55.4&51.5
& 66.2&60.7&68.4
\\
Qwen2.5-VL-72B &  67.5 
&67.0 &63.1&55.0 
&68.9 &63.8&57.9 
&76.7&77.1&66.8 
&72.7&71.5&68.0 
&71.2&67.3&68.7 
&69.3&65.1&65.4 
\\
InternVL2.5-MPO & 65.2  
&55.6 &50.9&43.2 
&64.6 &59.4&53.7 
&78.4&73.1&60.5 
&73.9&70.5&63.6 
&72.9&73.6&67.5 
&76.2&68.3&67.3 
\\
InternVL2.5-78B & 64.6 
&51.8 &48.5&41.1 
&60.9 &55.6&49.7 
&78.2&75.9&59.8 
&74.0&70.2&65.5 
&74.9&72.8&69.0 
&79.1&68.6&66.3 
\\
QVQ-72B &  64.4 
&60.9 & 62.3&54.3 
&60.6 &68.8&68.7 
&70.6&68.7&62.1 
&65.3&63.9&58.2 
&64.8&66.8&65.9 
&66.7&62.9&67.4 
\\
Qwen2.5-VL-7B & 60.3  
&59.7 &56.2&46.7 
&57.2 &56.2&46.7 
&66.2&67.6&56.9 
&64.2&62.3&59.8 
&61.7&63.0&67.9 
&63.6&64.6&62.0 
\\
InternVL2.5-8B & 54.6 
&46.1 &40.5&35.9 
&51.3 &45.0&38.8 
&63.9&65.3&51.1 
&59.4&56.7&54.3 
&64.3&58.1&58.5 
&73.5&62.1&57.4 
\\
Qwen2-VL-7B &  45.4 
&42.3 &37.9&31.6 
&45.7 &39.1&33.8 
&53.0&50.4&41.3 
&52.1&46.8&49.7 
&43.0&47.8&41.5 
&55.4&51.9&53.3 
\\
\bottomrule
    \end{tabularx}
  \label{tab:perf_tab1}%
\end{table*}%

\begin{figure*}[h]
    \centering
    \centering
    \includegraphics[trim=0.5cm 0.5cm 0.5cm 0.7cm, width=0.99\textwidth]{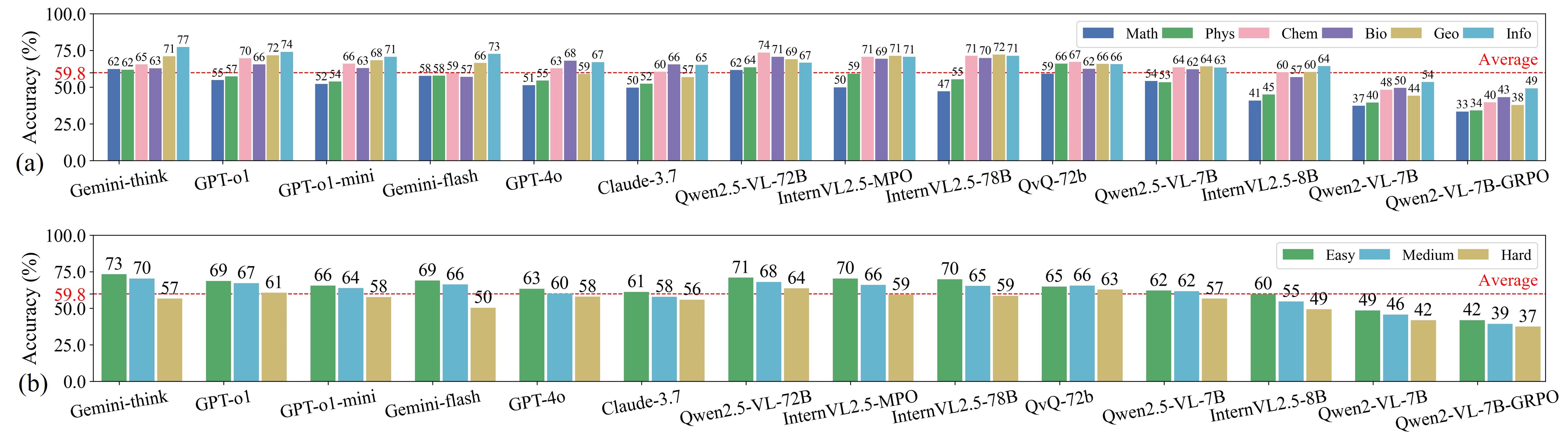}
    \caption{Comparison of average model accuracy with respect to (a) six disciplines and (b) exam difficulty levels}
    \label{fig:sub_diff}
\end{figure*}

\noindent\textbf{Data Processing.}
Following screening, we applied rule-based parsing to convert each exam instance into a uniform structured data format and converted into JSON and TSV files. To ensure linguistic and formatting consistency the GPT-4o is used to translate Chinese text into English and verified by domain experts for technical accuracy. Data experts then inspected the processed dataset for translation fidelity, content completeness, unit and encoding consistency, equation formatting, question categorization, and overall compliance with formatting standards. Rules were established to automatically remove instances that failed to meet these criteria, yielding a final dataset of 141.3K exam instances. Each instance includes the following data fields: difficulty level, exam year, question form, question, answer, text, image, grade level, curriculum, topic, knowledge points, and answer explanation.


\noindent\textbf{Knowledge Taxtonomy.}
We constructed a six-layer knowledge taxonomy from the processed data: Level 1 – disciplines, Level 2 – grade levels, Level 3 – subfields, Level 4 – curriculum, Level 5 – topics, and Level 6 – knowledge points. The benchmark covers six subjects: mathematics, physics, chemistry, biology, geography, and information science. Each subject includes middle school (K7–K9) and high school (K10–K12) grades, with mathematics additionally covering primary school (K0–K6). Data experts defined six subfields per subject, and GPT-4o was used to map 90 curricula to 36 subfields, followed by manual inspection. The complete list of subfields are shown in Fig. \ref{fig:data}. Each processed question is linked to this taxonomy, enabling structured, fine-grained knowledge representation.  The statistics is provide in Table. \ref{tab:dataset_stats}, and the distribution of Level 3–6 knowledge taxonomy is presented in the Supplementary Material.

\noindent\textbf{Data Statistics.} The statistics of MDK12-Bench is summarized in Table~\ref{tab:dataset_stats}, the full set of our benchmark \textbf{MDK12-Full} comprises 141,320 unique exam instances, including 77,857 (55.1\%) text-only and 63,463 (44.9\%) text–image pairs, totaling 105,218 images. Covering a 12-year span (2016–2025), it includes five question formats: single-choice, multiple-choice, fill-in-the-blank, true/false, and open-ended. To support lightweight evaluation, we introduce \textbf{MDK12-Mini}, consisting of 10\% of MDK12-Full uniformly sampled across easy, medium, and hard levels. Knowledge points are uniformly sampled to ensure each subset instance is linked to at least one unique knowledge point. In the Supplementary Material, we provide evidence demonstrating that MDK12-Mini yields evaluation results comparable to those of MDK12-Full.


\section{MLLM Evaluation} \label{sec:mllm_eval}

\subsection{Evaluation Methods}

As illustrated in Fig.~\ref{fig:dynamic_eva}, the evaluation is based on answer matching which comprises three steps: (1) Input question into the MLLMs to generate an answer; (2) GPT-4o is prompt to parse and extract the MLLM response as its final answer; and (3) MLLM answer is compared with the ground truth answer. compared against the ground truth. Exact matches score 1.0, while partial matches are graded by GPT-as-Judge using predefined rules (e.g., 0.5 for one of two filled blanks, m/n for m correct out of n choices). In \textbf{Standard} evaluation, the MLLM is given the original question and evaluated against its ground truth answer. \textbf{Dynamic} evaluation transforms both the question and its ground truth, while \textbf{KP-RAG} evaluation enriches the question with relevant knowledge points, prompting the model to elaborate on these points and answer using both the question and expanded knowledge.






\begin{table*}[h]
\centering
\small
\setlength{\tabcolsep}{3.8pt}
\renewcommand{\arraystretch}{1}
\caption{Accuracy of MLLMs on standard vs. dynamic evaluation; $\Delta$ shows their difference. \textbf{Best} and \underline{second best} highlighted.}
\begin{tabular}{lcccccccccccc}
\toprule
\multicolumn{1}{c}{\multirow{2}{*}{\textbf{Model}}} 
& \multicolumn{3}{c}{\textbf{Overall}} 
& \multicolumn{3}{c}{\textbf{Easy}} 
& \multicolumn{3}{c}{\textbf{Medium}} 
& \multicolumn{3}{c}{\textbf{Hard}} \\
\cmidrule(lr){2-4}\cmidrule(lr){5-7}\cmidrule(lr){8-10}\cmidrule(lr){11-13}
& Standard & Dynamic & \textbf{$\Delta$}
& Standard & Dynamic & \textbf{$\Delta$}
& Standard & Dynamic & \textbf{$\Delta$}
& Standard & Dynamic & \textbf{$\Delta$} \\
\midrule
Gemini2-thinking 
& \textbf{58.1} & \underline{41.6} & 16.5
& \textbf{66.7} & \textbf{43.8} & 22.9
& \textbf{57.0} & \textbf{44.8} & 12.2
& \textbf{51.5} & \underline{36.2} & 15.3 \\
Gemini2-flash 
& \underline{56.4} & \textbf{47.0} & {9.4}
& \underline{66.6} & \underline{50.1} & {16.4}
& \underline{54.7} & \underline{46.1} & {8.6}
& \underline{48.9} & \textbf{44.5} & {4.4} \\
GPT-4o 
& 51.2 & 40.9 & 10.3
& 54.1 & 35.7 & 18.5
& 53.7 & 51.3 & 2.4
& 35.4 & 34.8 & 0.6 \\
Claude-3.7 
& 46.7 & 31.4 & 15.3
& 49.2 & 32.3 & 16.9
& 50.2 & 36.3 & 13.9
& 40.5 & 25.2 & 15.3 \\
InternVL2.5-8B 
& 41.7 & 26.1 & 15.6 
& 48.5 & 23.5 & 25.0 
& 44.1 & 27.5 & 16.6 
& 38.4 & 30.8 & 7.7 \\
Qwen2-VL-7B 
& 27.3 & 26.1 & 1.2 
& 31.8 & 34.6 & -2.8 
& 25.5 & 25.4 & 0.0 
& 25.6 & 20.5 & 5.0 \\
\bottomrule
\end{tabular}
    \label{tab:dynamic}
\end{table*}

\begin{figure*}[ht]
    \centering
    \centering
    \includegraphics[trim=0cm 0.5cm 0 0, width=0.99\textwidth]{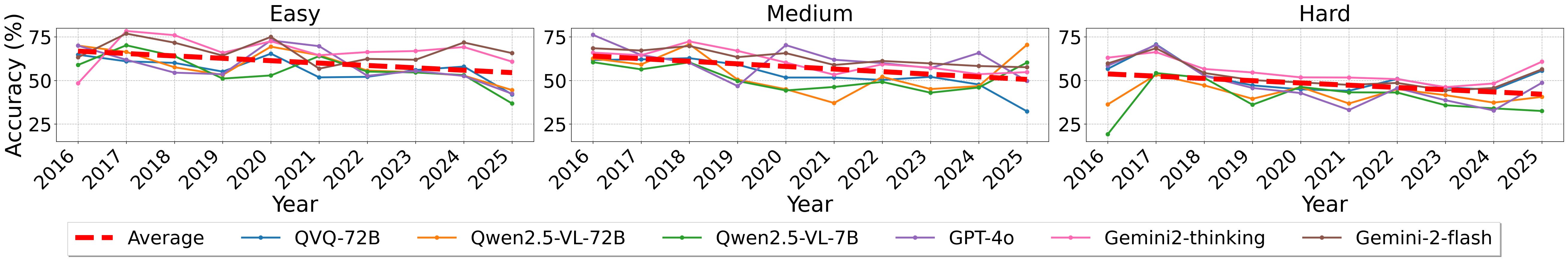}
     \caption{Model accuracy across exam years.}
    \label{fig:year_breakdown}
\end{figure*}
\begin{figure*}[h]
    \centering
    \centering
    \includegraphics[trim=0cm 0.5cm 0 0, width=0.99\textwidth]{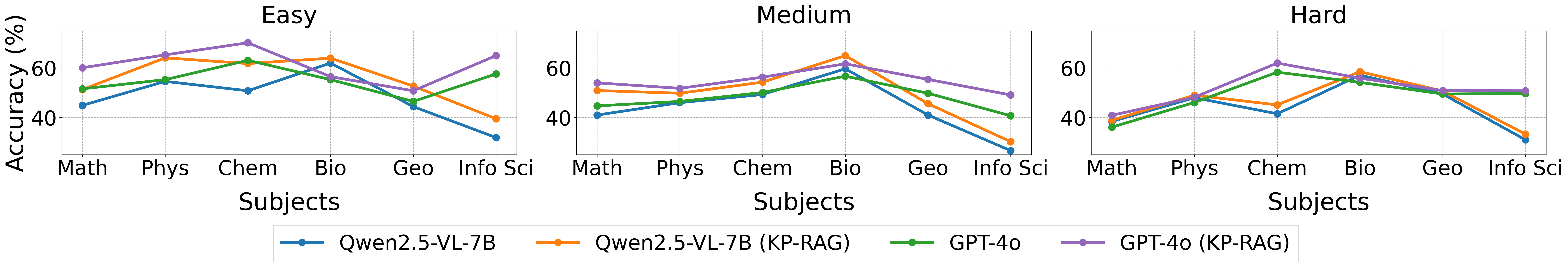}
    \caption{Comparison of model accuracy with vs. without KP-RAG.}
    \label{fig:kg}
    \vspace{-0.2em}
\end{figure*}

The dynamic evaluation framework introduces controlled perturbations to images and texts, including both questions and ground-truth answers. It creates new test samples during MLLM evaluation.  A GPT-based judge checks whether dynamic text or image alterations alter the meaning of the original question or answer and removes samples with significant changes. The dynamic transform include visual and textual transforms as detailed below. Details of the dynamic bootstrapping and evaluation algorithm, as well as the GPT-based judge prompt, are included in the Supplementary Material.


Image bootstrapping strategies apply spatial, color, and style variations to increase visual diversity and difficulty in visual recognition and reasoning. They preserve semantics while making the context unfamiliar to models. \textbf{Spatial.} We pad the original image with colors uniformly sampled from black, white, and grey. The padding width is proportional to the image dimension along each side, with the ratios uniformly sampled in the range between 10$\%$ and 20$\%$. The image padding allows the evaluation of a model’s ability to recognize layout changes and apply its structural knowledge to compare, contrast, and reason about image layout changes. \textbf{Color.} In this step, the colors of the original image were inverted. Salt-and-pepper noise of random noise density is also added. This transformation assesses the model’s ability to utilize its familiar visual knowledge to identify and reason about visual entities when image experience significant color distortions and random visual artifacts. \textbf{Style.} Using the Flux-Dev \cite{blackforestlabs_flux} model, we introduce subtle style variations without significantly altering its key visual elements and question semantics. This tests whether models can rely on its physical appearance understanding and knowledge to reconize and reason over unseen style variations.

We also introduce three textual bootstrapping methods to modify questions while preserving the answer's correctness.\textbf{Word Substitution.} We replace certain keywords with synonyms or contextually related expressions. This tests how well a model can maintain an accurate understanding when familiar terms are changed, thus assessing vocabulary sensitivity and semantic generalization. \textbf{Sentence Paraphrasing.} We rephrase entire sentences through variations in sentence structure, word order, or style. This checks whether a model can consistently capture the underlying meaning even when the surface form of the text is altered. \textbf{Question Type Conversion.} We convert a question from one format to another, such as turning a multiple-choice problem into a fill-in-the-blank.

\section{Experiments}
\label{sec:experiments}


\subsection{Baselines and Experimental Setup}
\label{subsec:baseline_setup}

We conduct systematic and fine-grained evaluation of a set of ten existing MLLMs from multiple dimensions. The baseline MLLMs including  \textbf{Proprietary} models: Gemini-2.0-flash-exp \cite{team2023gemini}, Gemini-2.0-flash-thinking-exp \cite{team2023gemini}, GPT-4o \cite{openai2024gpt4o}, GPT-o1-mini \cite{openai2024gpto1mini}, Claude-3.7-Sonnet \cite{claud3}. \textbf{Open-source} models: Qwen2.5-VL \cite{bai2025qwen2}, InternVL2.5 \cite{chen2024expandinginternvl2.5}, QVQ-72B-preview \cite{qvq-72b-preview}, InternVL2.5-78B-MPO \cite{wang2024mpo}.  All experiments are performed in a zero-shot setup and evaluated by accuracy metric, demonstrating MLLMs’ ability to generalize in multidisciplinary problem solving without few-shot prompting or model fine-tuning. In the following discussions, we also compare \textbf{Reasoning} models with their \textbf{Chat} counterparts, such as Gemini2-think vs. Gemini2-flash, GPT-o1 vs. GPT-4o, and InternVL2.5-MPO vs. InternVL2.5-78B.



\subsection{Cross-Discipline and Difficulty Results}
\label{subsec:main_table_results}

We present performance of baseline MLLMs across six disciplines on the easy, medium, and hard subsets of MDK12-Mini in Table~\ref{tab:perf_tab1}. \textbf{Cross-Disciplines} performance is shown in Fig.~\ref{fig:sub_diff}(a), illustrates the average accuracy of each model. Models consistently perform worse in Mathematics and Physics, with scores 7.6\% lower than the overall average of 59.8\%. In contrast, Chemistry, Biology, Geography, and Information Science achieve an average score that 3.7\% higher than the overall average. \textbf{Cross-Difficulty} performance is presented in Fig.\ref{fig:sub_diff}(b), which shows the average accuracy of each model across the three difficulty levels. All models show decreased accuracy on harder exam questions, with an average drop of 8.3\% compared to easier ones. Larger models consistently outperform smaller ones across disciplines and difficulty levels. Separately, reasoning-oriented models also achieve higher accuracy than their chat-focused counterparts, a trend more evident in the Gemini and GPT series than in the Intern series. The evaluation on MDK-Full is provided in the Supplementary Material and shows results consistent with those of MDK-Mini.


\subsection{Cross-Year Evaluation Results}
 A year-by-year accuracy breakdown across difficulty levels is shown in Fig.~\ref{fig:year_breakdown}, which showcases the temporal shifts in model performance relative to exam year. While accuracy naturally declines with increasing difficulty, we observe a further performance drop on newer exams across all difficulty levels, with accuracy gaps of 12.3\%, 13.4\%, and 11.6\% between the oldest and newest exams for easy, medium, and hard levels, respectively. This temporal trend may stem from distributional shifts where newer exams introduce updated novel concepts or rephrased questions that differ from the training data of models. As a result, models face reduced familiarity and limited exposure, leading to lower accuracy even at similar difficulty levels.

\subsection{Dynamic Evaluation Results}
\label{subsec:dynamic_eval}



We sampled 50\% of MDK12-Mini’s multimodal instances (695 easy, 818 medium, 1124 hard) as the standard set and generated a dynamic query set using three textual and three visual bootstrapping methods. Table~\ref{tab:dynamic} shows that dynamic evaluation reduces model performance by an average of 13.7\%, indicating their generalization limitations. Despite strong baseline performance, leading models especially reasoning model (e.g., Gemini-think) are more sensitive to contextual shifts possibly due to their stronger context-aware capability, more complex reasoning chain, or overfitting to massive static pretraining corpora, which dynamic perturbations easily disrupt.  Ablation studies for individual transformations are provided in the Supplementary Material.

\subsection{KP-RAG Evaluation Results}
We compare the accuracy of Qwen2.5-VL-7B and GPT-4o with and without knowledge-point referenced generation (KP-RAG) in Fig.~\ref{fig:kg}. It is observed that incorporating KP-RAG improves model accuracy by an average of 6.9\%, 6.0\%, and 2.1\% on easy, medium, and hard exams, respectively. The larger gains on easy and medium exams likely arise because these questions are less reasoning-intensive and more knowledge-retrieval driven, allowing explicit knowledge-point augmentation to benefit the model. In contrast, harder exams often require multi-step reasoning, abstract problem-solving, or cross-knowledge integration, where simply adding related knowledge points offers limited improvement.

\subsection{Error Analysis}
\begin{figure}[t]
    \centering
    \includegraphics[trim=0.5 0.5cm 0 0, width=0.8\linewidth]{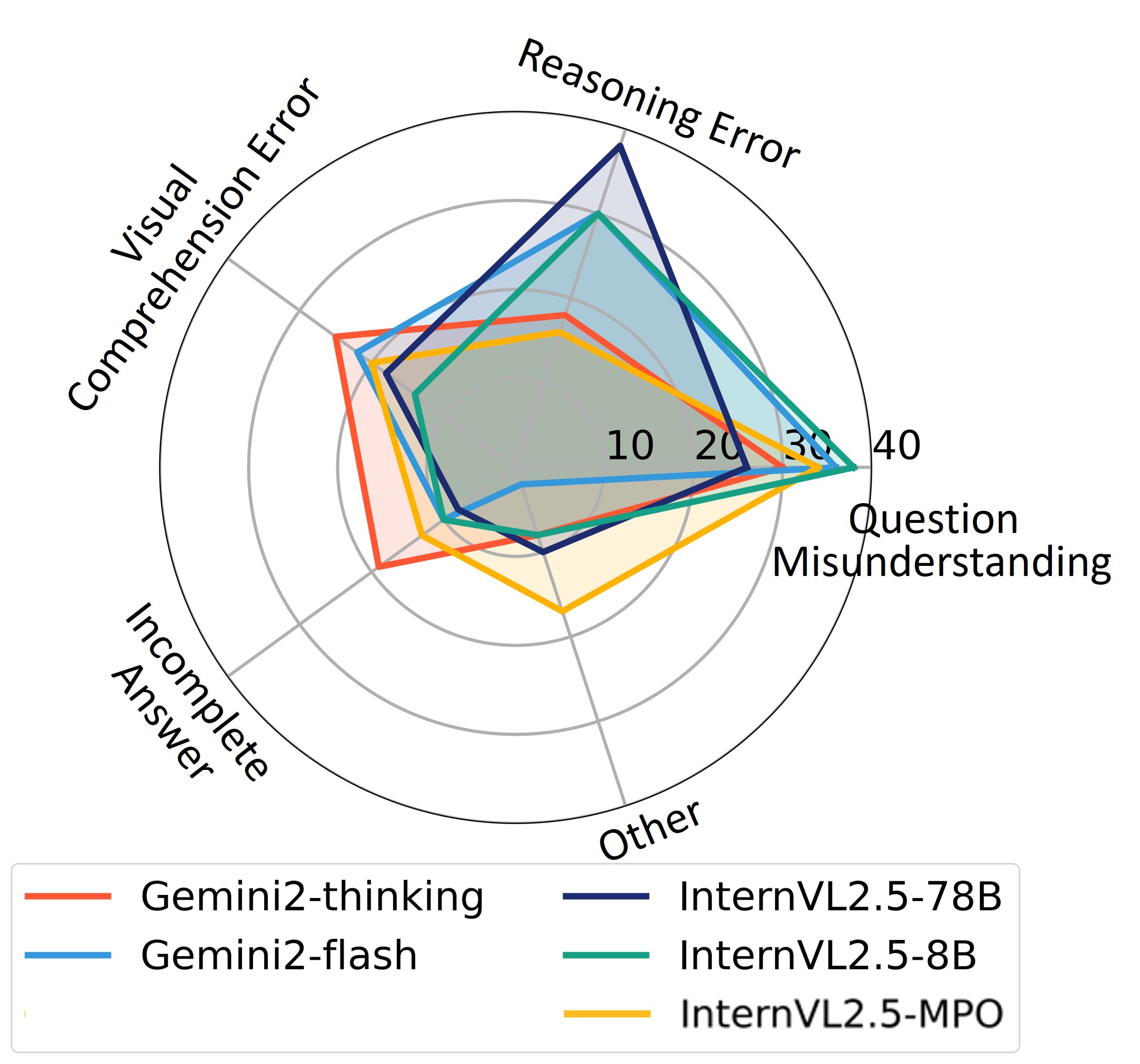}
    \caption{Error analysis of five representative models.}
    \label{fig:error_analysis}
    \vspace{-0.2cm}
\end{figure}

We analyze 100 sampled errors from five models: Gemini2-thinking, Gemini2-flash, InternVL2.5-MPO, InternVL2.5-78B, and InternVL2.5-8B, and categorized them into five types (Fig.~\ref{fig:error_analysis}): Question Misunderstanding, Reasoning Error, Visual Comprehension Error, Incomplete Answers, and Other Errors. Reasoning models (Gemini-think and InternVL2.5-MPO) reduce reasoning errors by 12\% and 22\% compared to their chat counterparts (Gemini2-flash and InternVL2.5-78B), likely due to RL-based reasoning-oriented post-training, but exhibit 2\% higher visual errors and more incomplete answers (12\%, 5\%) due to vision encoder limits and the 2048-token cutoff. Larger models (e.g., InternVL2.5-78B) reduce visual and other errors by 3\% and 1\% compared to InternVL2.5-8B but show no improvement in reasoning, question understanding, or incomplete answers. This suggests that scaling mainly enhances perception through richer multimodal representations but little to improve reasoning accuracy, prompt interpretation, instruction-following, or token-length limitations.

\section{Conclusion}
We present MDK12-Bench, a comprehensive multimodal benchmark for evaluating problem-solving intelligence of MLLMs across diverse disciplines and dimensions based on real-world K-12 exams. Spanning 141K questions, MDK12-Bench addresses key limitations of existing benchmarks, including limited scale, lack of fine-grained annotations, and unstructured knowledge representation. To ensure scalable, objective, and long-term evaluation, we propose a dynamic framework that applies diverse textual and visual bootstrapping strategies to rigorously assess model generalization and mitigate data contamination. Experimental results reveal significant limitations of current state-of-the-art MLLMs, including high sensitivity to contextual changes, poor generalization to novel and complex tasks, and limited benefits of knowledge augmentation to solve reasoning-intensive problems. These findings affirm the role of MDK-benchmark as an essential foundation for diagnosing strengths and limitations of current models and for steering the development of robust and generalizable multimodal intelligence through improved adaptability, reasoning, and knowledge integration.

\twocolumn[{%
 \centering
 \Large \textbf{MDK12-Bench: A Comprehensive Evaluation of \\ Multimodal Large Language Models on Multidisciplinary Exams} \\[0.3em]Supplementary Material\\[1.5em]
}]

\appendix
\label{sec:appendix}

\setcounter{figure}{9}
\setcounter{table}{5}
\setcounter{equation}{6}

\maketitle
In this supplementary material, we present additional information, discussions, and results that support the main text, organized as follows:

\begin{figure*}[h]
    \centering
    \centering
    \includegraphics[trim=0.5cm 0.5cm 0.5cm 0.7cm, width=0.99\textwidth]{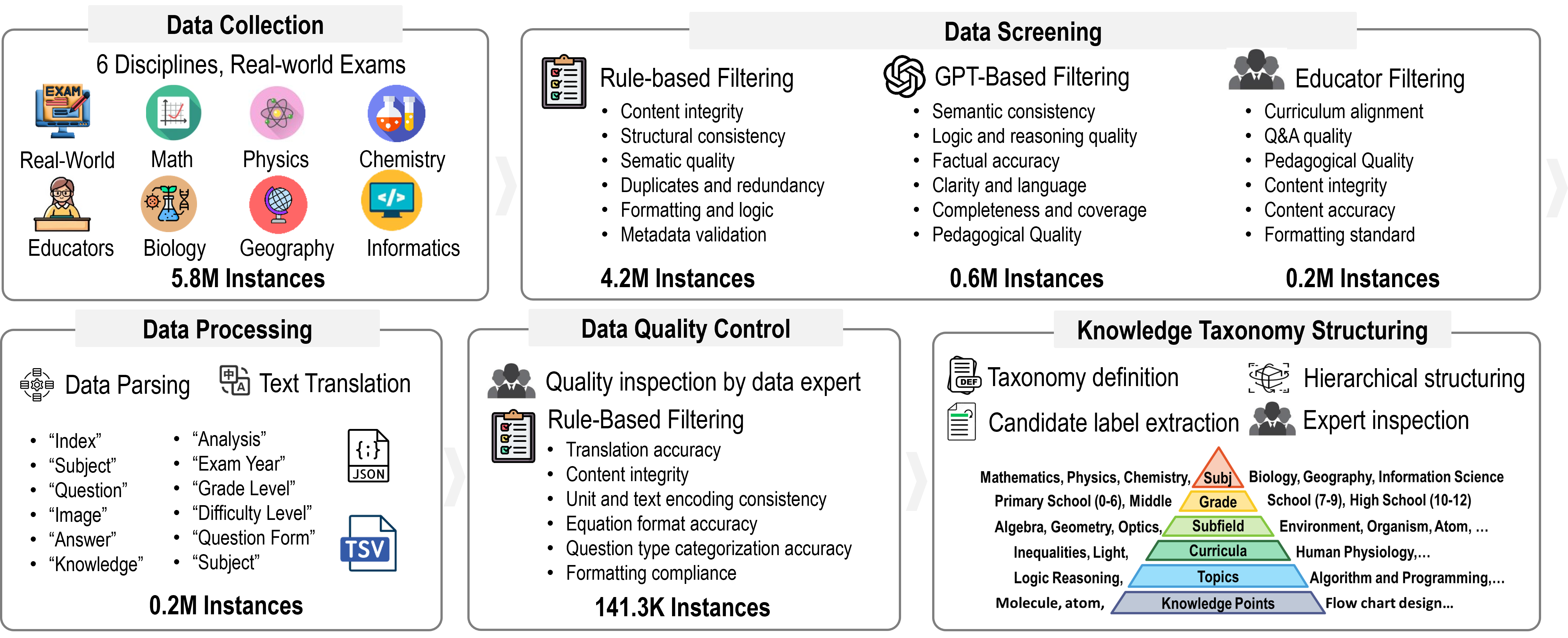}
    \caption{Pipeline of data curation}
    \label{fig:data_pipeline}
\end{figure*}

\begin{algorithm}[ht]
\caption{Dynamic Bootstrapping \& Evaluation}
\label{alg:example}
\begin{algorithmic}[1]
\Require 
  \(\mathcal{S} = \{(Q,I,A)\}\)\Comment{Original dataset: each sample has question \(Q\), image \(I\), and ground-truth \(A\)}
  \Statex \(\mathcal{T}_{\mathrm{text}}\)\Comment{Set of textual transformations (e.g., word substitution, paraphrasing and type permuting)}
  \Statex \(\mathcal{T}_{\mathrm{vis}}\)\Comment{Set of visual transformations (e.g., expansion, color shift and style transfer)}
  \Statex \(M\)\Comment{The MLLM to be evaluated}
\Ensure 
  \(\text{Scores}\)\Comment{Final evaluated scores}

\State \(\mathcal{S}_{aug} \gets \varnothing\) \Comment{Augmented dataset (initially empty)}

\For{each \((Q,I,A) \in \mathcal{S}\)}
  \State \(\mathcal{Q'} \gets \varnothing,\; \mathcal{I'} \gets \varnothing\) \Comment{Sets for valid textual/visual variants}
  \For{each \(t \in \mathcal{T}_{\mathrm{text}}\)}
    \State \(Q_t \gets \mathrm{TextTransform}(Q, t)\) \Comment{Transform question \(Q\) with method \(t\)}
    \If{\(\mathrm{Judge}(Q_t, A) = \mathrm{True}\)}
      \State \(\mathcal{Q'} \gets \mathcal{Q'} \cup \{Q_t\}\)
    \EndIf
  \EndFor

  \For{each \(v \in \mathcal{T}_{\mathrm{vis}}\)}
    \State \(I_v \gets \mathrm{VisTransform}(I, v)\) \Comment{Transform image \(I\) with method \(v\)}
    \If{\(\mathrm{Judge}(I_v, A) = \mathrm{True}\)}
      \State \(\mathcal{I'} \gets \mathcal{I'} \cup \{I_v\}\)
    \EndIf
  \EndFor

  \State \(\mathcal{Q'} \gets \mathcal{Q'} \cup \{Q\}\) \Comment{Include original question}
  \State \(\mathcal{I'} \gets \mathcal{I'} \cup \{I\}\) \Comment{Include original image}

  \For{each \(Q^* \in \mathcal{Q'}\)}
    \For{each \(I^* \in \mathcal{I'}\)}
      \State \(\mathcal{S}_{aug} \gets \mathcal{S}_{aug} \cup \{(Q^*, I^*, A)\}\) \Comment{Construct new sample}
    \EndFor
  \EndFor
\EndFor

\State \(\text{scores} \gets \varnothing\) \Comment{List of model scores}
\For{each \((Q_d, I_d, A) \in \mathcal{S}_{aug}\)}
  \State \(r \gets M(Q_d, I_d)\) \Comment{Model output (raw response)}
  \State \(a \gets \mathrm{Parse}(r)\) \Comment{Extract final predicted answer from \(r\)}
  \If{\(a = A\)}
    \State \(s \gets 1\) \Comment{Exact match}
  \Else
    \State \(s \gets \mathrm{Partial}(a,A)\) \Comment{Partial-credit scoring}
  \EndIf
  \State \(\text{scores} \gets \text{scores} \cup \{s\}\)
\EndFor
\State \Return \(\text{scores}\)
\end{algorithmic}
\end{algorithm}

\noindent
\begin{itemize}





\item Data Curation and Taxonomy: Detailed description of the MDK12-Bench data collection process and its structured knowledge taxonomy.

\item Evaluation Pipeline: Expanded explanation of the evaluation process, including answer matching, dynamic evaluation, and knowledge-referenced evaluation, highlighting the novel approaches used for fair and consistent performance assessment.

\item Additional Experiments: Extended experiments validating model performance under various scenarios, including ablation studies on dynamic shift strategies, comparative results between MDK-Full and MDK-Mini, and bilingual evaluation outcomes, providing deeper insights into MLLMs’ strengths and limitations.

\item Case Studies: Illustrative examples showcasing model behaviors along with analyses of different error types.

\item Future Work: Discussion of potential directions for advancing multimodal large language model evaluation and robustness.

\end{itemize}

\section{Details of MDK12-Bench and Data Curation}
\label{sec:data_curation}

\subsection{Data Curation}
To build MDK12-Bench, we curated 5.8M multimodal K–12 exam instances from open-access repositories spanning six disciplines and multiple grades, curricula, and difficulty levels. The data underwent a four-stage pipeline illustrated in Fig.\ref{fig:data_pipeline} and summarized in Table.\ref{tab:data_screening}, which lists the filtering rules and the remaining number of instances after each stage. The first three stages comprise pre-processing and parsing: (1) Rule-based filtering automatically removed low-quality or irrelevant questions based on text–image correspondence, formatting, semantic coherence, duplication, and logical soundness, reducing the dataset to 4.2M instances. (2) GPT-based filtering used GPT-4o to assess semantic consistency, reasoning validity, factual correctness, and language clarity, narrowing the data to 0.6M. (3) Manual educator review ensured curriculum alignment, question–answer correctness, and pedagogical quality, producing 0.2M high-quality instances.

Following screening, a post-processing stage standardized the data into JSON and TSV formats. GPT-4o translated Chinese text into English, with domain experts verifying technical accuracy. Data experts performed final checks for translation fidelity, content completeness, unit and encoding consistency, equation formatting, question categorization, and compliance with formatting standards. Instances failing these checks were automatically removed, yielding a final benchmark of 141.3K exam instances, each annotated with difficulty level, exam year, question format, answer, explanation, grade level, curriculum, topic, and knowledge points.

\subsection{Knowledge Taxonomy and Statistics}
\textbf{Knowledge Taxonomy Construction}. As illustrated in Fig.\ref{fig:data_pipeline}, we constructed a six-layer knowledge taxonomy from the processed data: Level 1 – disciplines, Level 2 – grade levels, Level 3 – subfields, Level 4 – curriculum, Level 5 – topics, and Level 6 – knowledge points. The benchmark covers six subjects: mathematics, physics, chemistry, biology, geography, and information science. Each subject includes middle school (K7–K9) and high school (K10–K12) grades, with mathematics additionally covering primary school (K0–K6). The subfields covered by each discipline are shown in Table~\ref{tab:discipline_subfields}. These subfields span core topics within mathematics, physics, chemistry, biology, geography, and information science, ensuring thorough coverage of K–12 educational curricula. We construct a knowledge point mapping to organize the hierarchical taxonomy of multidisciplinary K–12 exams. We recursively traverse each course’s tree-structured knowledge representation, extracting node names at every level and ensuring uniqueness across the hierarchy. This process builds a multi-level mapping from broad discipline and subfield (level 1-3) to increasingly fine-grained concepts (levels 4-6). Parent–child relationships are explicitly recorded to preserve the full dependency structure, enabling us to link each knowledge point to its ancestors. The resulting taxonomy captures over six thousand unique knowledge points with their hierarchical relations, supporting fine-grained reasoning evaluation and structured knowledge-augmented generation.

To further illustrate the benchmark’s knowledge statistics, Fig.\ref{fig:stats_middle_school} depicts the distribution of knowledge points for primary and middle school levels (K0–K9), while Fig.\ref{fig:stats_highschool} presents the corresponding distribution for high school levels (K10–K12). The knowledge distributions in both figures exhibit a pronounced long-tailed pattern, where a small subset of high-frequency knowledge points accounts for a large proportion of questions, while numerous low-frequency points appear sparsely.

\textbf{Knowledge Point Distribution Across Grade Levels}. A quantitative analysis shows that the top 10\% most frequent knowledge points contribute to approximately 70–80\% of all exam instances, whereas nearly 40\% of knowledge points appear in fewer than 0.1\% of instances each. This long-tailed behavior reflects real-world exam design: foundational concepts (e.g., basic arithmetic or Newtonian mechanics) are repeatedly assessed, whereas specialized or advanced topics occur less frequently. Such a distribution introduces challenges for model training and evaluation, as models must demonstrate not only strong performance on common knowledge areas but also robust generalization to rare, less-represented concepts. MDK12-Bench preserves this natural imbalance to better mirror authentic educational settings and to test the adaptability and coverage of multimodal large language models.

\textbf{Knowledge Point Ranking by Model Accuracy}. Fig.~\ref{fig:eva_acc_by_knowledge} ranks knowledge points based on problem-solving accuracy achieved by Gemini2-think. We report the top 5\% of knowledge points where the model performs best and the bottom 25\% where performance is weakest. The results show that high-performing points largely correspond to frequently tested foundational topics, while low-performing points are concentrated in advanced or specialized concepts such as complex geometry and biochemical processes. This ranking enables fine-grained identification of knowledge gaps, providing actionable insights for improving model reasoning capabilities and guiding targeted dataset augmentation.

\begin{figure}[t]
    \centering
    \centering
    \includegraphics[trim=0.5cm 0.5cm 0.5cm 0.7cm, width=0.45\textwidth]{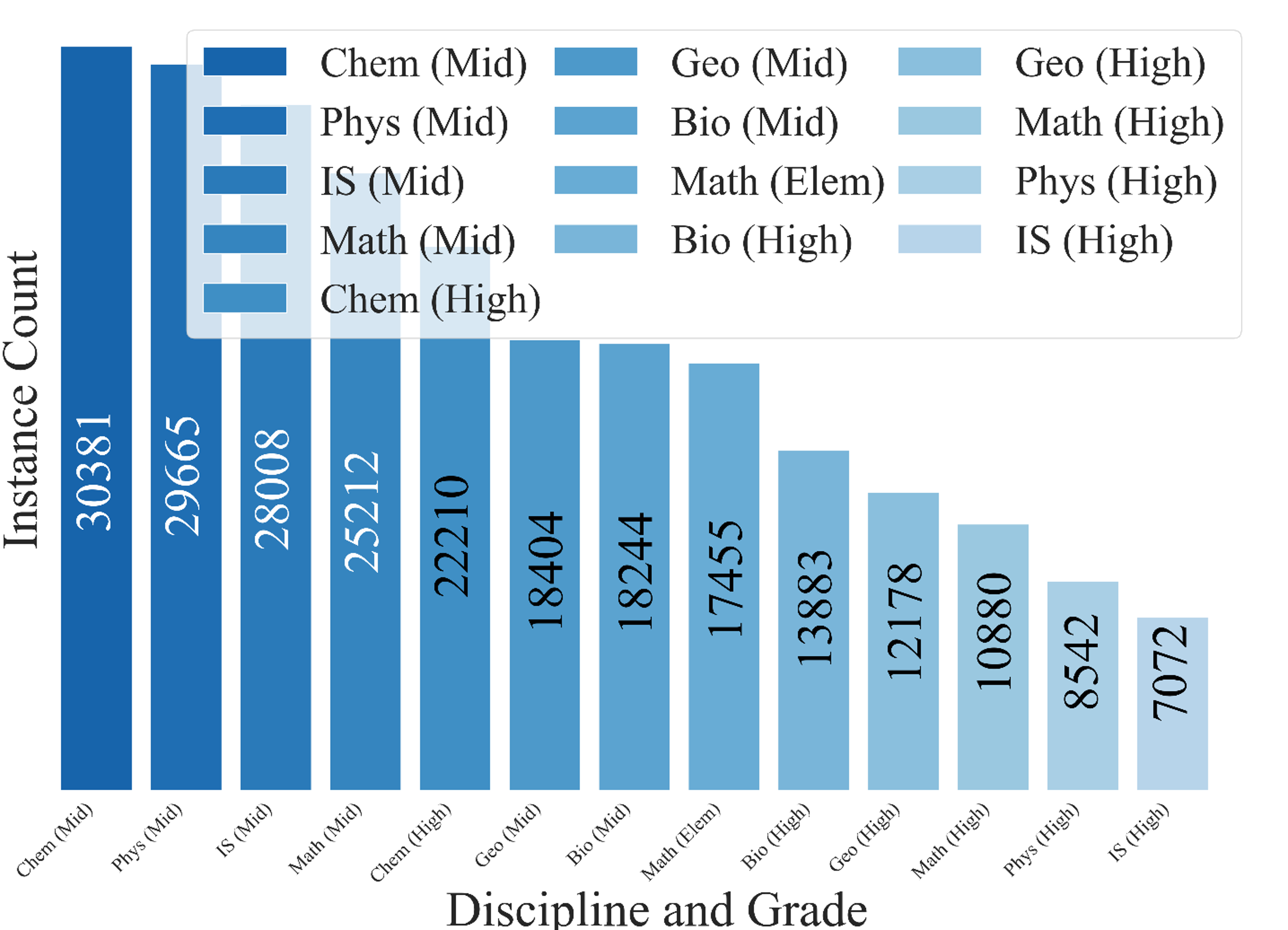}
    \caption{Data distribution over discipline and grade}
    \label{fig:stats_instnum}
\end{figure}

\begin{figure*}[h]
    \centering
    \centering
    \includegraphics[trim=0.5cm 0.5cm 0.5cm 0.7cm, width=0.99\textwidth]{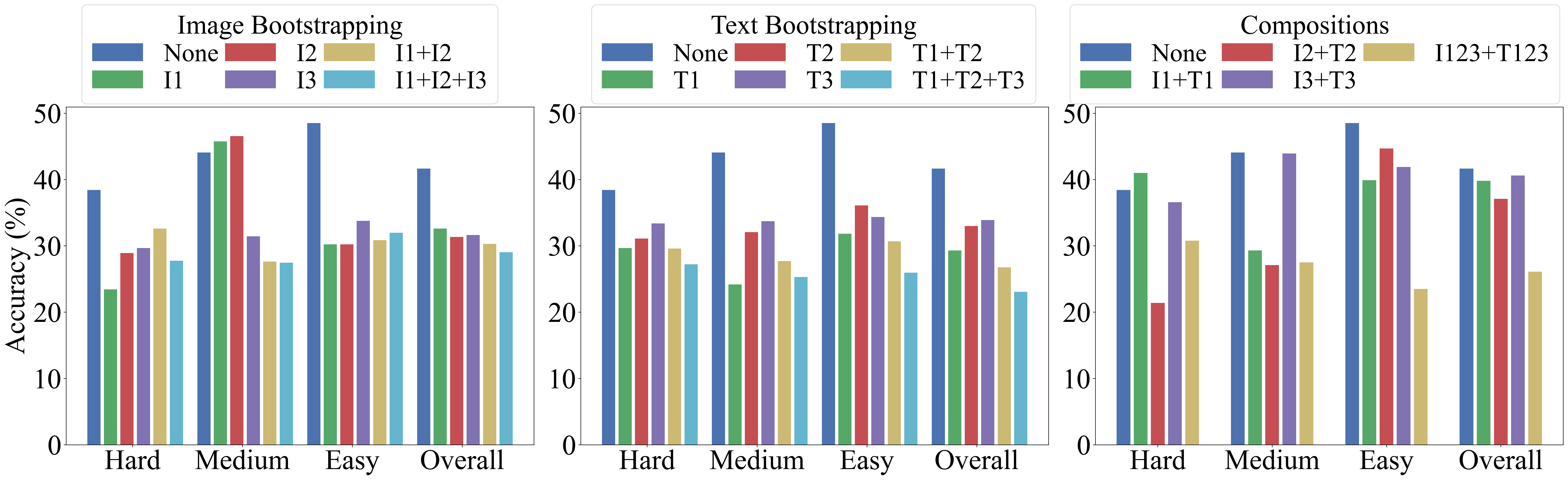}
    \caption{Ablation study of dynamic evaluation}
    \label{fig:dyn_ablation}
\end{figure*}

\section{Details of Evaluation Methods}
\label{sec:eva}
\subsection{Answer Matching}
Model performance is evaluated by sequentially feeding each sample from the augmented dataset into the target model $M$, collecting its raw responses, and extracting final predicted answers through a dedicated parsing function. The evaluation framework supports both exact matching and GPT-Judge scoring to capture model accuracy beyond strict binary correctness. Exact Matching relies on strict string comparison, is fast and deterministic without requiring external APIs, and is best suited for controlled answer formats with minimal variation. In contrast, GPT-Judge semantically evaluates whether predictions and ground truth share the same meaning, offering more intelligent handling of complex, free-form, descriptive, multilingual, or noisy predictions, albeit at the cost of slower performance and API dependency. Representative examples comparing these scoring methods are shown in Table~\ref{tab:gpt_vs_exact}.

 
\subsection{Dynamic Evaluation}
The evaluation pipeline for our proposed approach is outlined in Algorithm \ref{alg:example}, which systematically combines dynamic bootstrapping and model evaluation. The process begins with an original dataset $\mathcal{S}$ containing samples of question-image-answer triplets $(Q,I,A)$. Two sets of transformations, $\mathcal{T}_{\mathrm{text}}$ for textual augmentation and $\mathcal{T}_{\mathrm{vis}}$ for visual augmentation, are applied to generate diverse variants of the original samples. These transformations include methods such as word substitution, paraphrasing, question type transfer and image style transfer.

For each input sample, all valid textual and visual variants are generated using transformation methods that maintain the original answer's correctness, as verified by a judge function. The resulting augmented dataset $\mathcal{S}_{aug}$ is constructed by pairing all valid question and image variants with their corresponding (transferred) ground-truth answers.

The prompts for the four distinct question transformation types are illustrated in Figure~\ref{fig:dyn_trans_mc} (multiple-choice), Figure~\ref{fig:dyn_trans_fill} (fill-in-the-blank), Figure~\ref{fig:dyn_trans_open} (open-ended), Figure~\ref{fig:dyn_trans_truefalse} (true/false), and Figure~\ref{fig:dyn_trans_t2} (sentence paraphrasing). Specifically, for the Sentence Paraphrasing (T2) transformation, we provide detailed examples of rephrasing techniques using a few-shot prompting approach. These examples demonstrate how the model can generate syntactically diverse yet semantically equivalent reformulations of the original questions. By leveraging these structured prompts, the model can produce a wider range of question variants, contributing to a more diverse and robust augmented dataset. This approach not only enhances the model's adaptability to different question formats but also strengthens its reasoning and comprehension capabilities across varied academic subjects and difficulty levels. The prompt to check the semantic correctness of dynamically transformed question and answers Fig. \ref{fig:prompt_qa_judge}.

\subsection{KP-RAG Evaluation}
\begin{figure*}[h]
    \centering
    \centering
    \includegraphics[trim=0.5cm 0.5cm 0.5cm 0.7cm, width=0.7\textwidth]{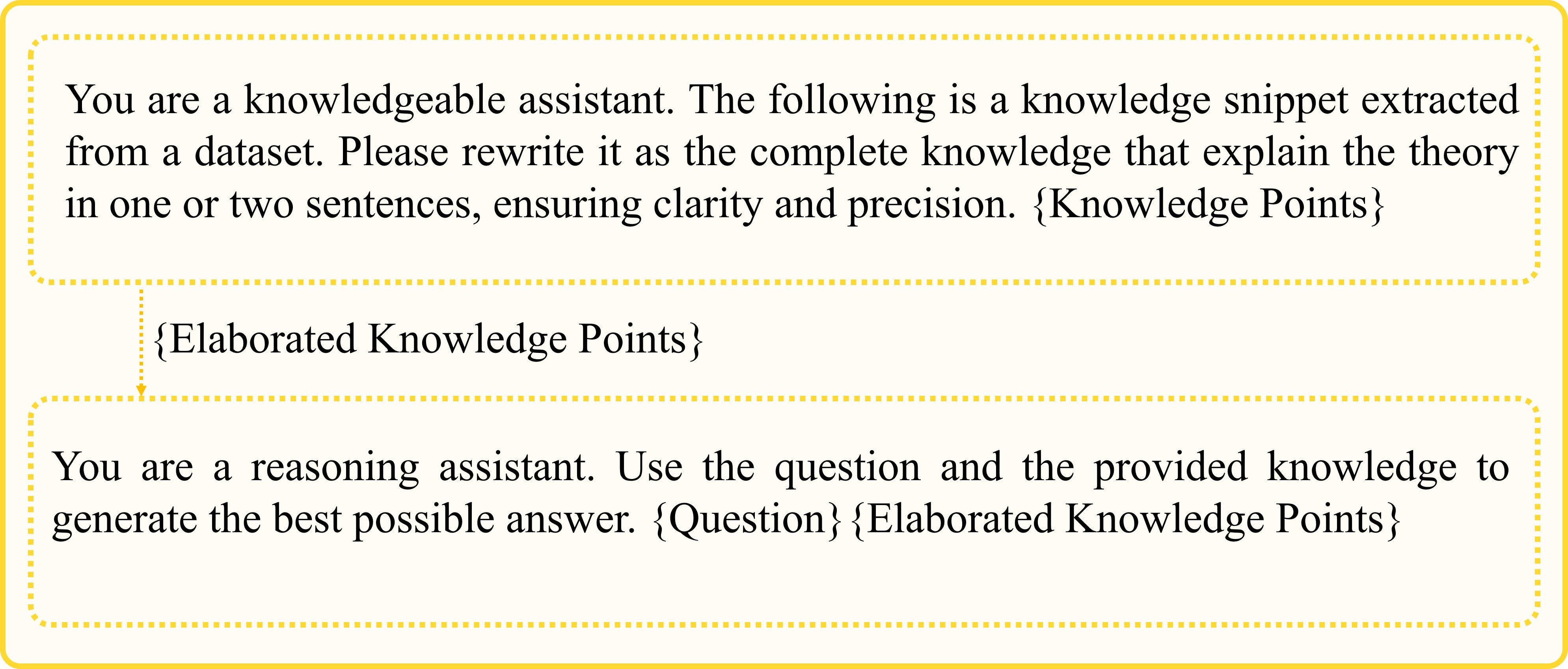}
    \caption{Prompt for KP-RAG.}
    \label{fig:stats_prompt_kprag}
\end{figure*}

Knowledge-point (KP) reference-augmented generation (KP-RAG) evaluation enriches each question with a set of relevant, fine-grained knowledge points retrieved from our hierarchical taxonomy. These knowledge points are appended to the original question and provided as structured hints, prompting the model not only to focus on the immediate context but also to incorporate domain-specific background knowledge during reasoning. By explicitly exposing the underlying concepts required to solve a question, KP-RAG encourages models to explain intermediate steps, elaborate on conceptual relationships, and produce more complete and interpretable answers. This evaluation setting is particularly useful for diagnosing whether a model can leverage structured prior knowledge to improve reasoning accuracy, handle multi-hop dependencies, and generalize to unseen questions that require understanding beyond surface-level text patterns. The prompt for knowledge reference generation is shown in Fig. \ref{fig:stats_prompt_kprag}.

\section{Additional Experiments} \label{sec:additional_experiments} 

\subsection{Baseline Models}
To comprehensively evaluate MDK12-Bench, we benchmarked a diverse set of multimodal large language models (MLLMs) encompassing both reasoning-focused and general-purpose chat models. The reasoning models are specifically optimized for deliberate, multi-step problem-solving and enhanced reasoning accuracy, whereas the chat models are designed for balanced, fast inference and general multimodal understanding. This selection includes state-of-the-art proprietary systems and competitive open-source models. Below, we detail each baseline model used in our experiments.

\begin{itemize}
\item \textbf{Reasoning Models}
\begin{itemize}
\item \textbf{Gemini-2-think}: Gemini-2.0-flash-thinking-exp: A reasoning-oriented variant of Gemini-2.0-flash designed for more deliberate and step-by-step problem-solving, improving performance on reasoning-intensive tasks \cite{team2023gemini}.

\item \textbf{GPT-o1-mini}: A smaller reasoning-focused OpenAI model optimized for multi-step thinking and efficient reasoning, offering improved performance on structured problem-solving while being lightweight \cite{openai2024gpto1mini}.

\item \textbf{Claude}: Claude-3.7-Sonnet is Anthropic’s latest multimodal model designed for safe and accurate reasoning. Claude-3.7-Sonnet handles complex instructions and multi-disciplinary problem-solving with strong text-image understanding \cite{claud3}.

\item \textbf{InternVL2.5-MPO}: InternVL2.5-78B-MPO is a reasoning-augmented version of InternVL2.5 with 78B parameters, leveraging multi-process optimization (MPO) for improved step-by-step reasoning and long-context understanding \cite{wang2024mpo}.

\item \textbf{Qwen2.5‑VL‑72B}: Qwen2.5‑VL‑72B – The largest model in the Qwen2.5-VL family, noted for state-of-the-art vision-language reasoning. 

\item \textbf{QVQ-72B-preview}: A large-scale, open-source preview model featuring 72B parameters, aimed at bridging reasoning and multimodal perception for challenging academic and real-world tasks \cite{qvq-72b-preview}
\end{itemize}

\item \textbf{Chat Models}
\begin{itemize}
\item \textbf{Gemini-2-flash}: Gemini-2.0-flash-exp: A proprietary MLLM developed  optimized for fast inference and general-purpose multimodal reasoning across text and images \cite{team2023gemini}.

\item \textbf{GPT-4o} : OpenAI’s flagship multimodal model capable of processing text, images, and audio inputs. GPT-4o emphasizes balanced performance across disciplines with high generalization ability \cite{openai2024gpt4o}.

\item \textbf{Qwen2.5-VL-7B} and \noindent\textbf{Qwen2-VL-72B}: Two open-source multimodal models, supporting vision-language reasoning. It is trained on diverse datasets to handle text-rich and visually grounded problem-solving tasks \cite{bai2025qwen2}.

\item \textbf{InternVL2.5}: A high-performing open-source MLLM built to integrate advanced visual understanding with language reasoning, suitable for fine-grained image comprehension tasks \cite{chen2024expandinginternvl2.5}.
\end{itemize}
\end{itemize}

\newcolumntype{Y}{>{\centering\arraybackslash}X}
\begin{table*}[!htbp]
  \centering
  \small
  \begin{tabularx}{0.99\textwidth}{@{}
    >{\raggedright\arraybackslash}p{3.5cm} 
    *{10}{Y}                         
    @{ \hspace{0.08cm} }
  @{}}
    \toprule
    \multicolumn{1}{l}{\multirow{2}{*}{\hspace{-1.5mm}\textbf{Models}}} 
      & \multicolumn{10}{c}{\textbf{Years}} \\
    \cmidrule(lr){2-11}
& 2016 & 2017 & 2018 & 2019 & 2020 & 2021 & 2022 & 2023 & 2024 & 2025 \\
\midrule
\textbf{Easy} & & & & & & & & & & \\
QVQ-72B & 64.8 & 61.0 & 60.1 & 55.1 & 65.4 & 51.8 & 52.1 & 56.0 & 58.0 & 42.0 \\
Qwen2.5-VL-72B & 70.0 & 66.4 & 57.6 & 53.2 & 69.4 & 64.5 & 55.7 & 54.8 & 52.8 & 44.5 \\
GPT-4o & 70.0 & 61.9 & 54.5 & 53.6 & 73.0 & 69.7 & 52.7 & 55.5 & 52.6 & 42.4 \\
Gemini2-thinking & 48.4 & 78.4 & 76.0 & 65.9 & 72.5 & 64.5 & 66.3 & 66.9 & 69.2 & 60.8 \\
Gemini-2-flash & 63.3 & 76.9 & 71.7 & 64.3 & 75.0 & 56.8 & 62.4 & 61.9 & 71.8 & 65.7 \\
\midrule
\textbf{Medium} & & & & & & & & & & \\
QVQ-72B & 61.8 & 62.1 & 62.8 & 59.1 & 51.7 & 51.7 & 50.3 & 52.1 & 47.7 & 32.3 \\
Qwen2.5-VL-72B & 62.6 & 59.2 & 70.8 & 50.5 & 45.0 & 37.1 & 52.1 & 45.1 & 46.8 & 70.5 \\
GPT-4o & 76.2 & 64.6 & 60.1 & 46.8 & 70.3 & 61.9 & 60.0 & 57.1 & 65.8 & 49.7 \\
Gemini2-thinking & 66.0 & 64.5 & 72.4 & 67.0 & 60.3 & 53.3 & 59.3 & 57.4 & 53.6 & 54.8 \\
Gemini-2-flash & 68.5 & 67.2 & 69.8 & 63.4 & 65.7 & 58.9 & 61.2 & 59.8 & 58.3 & 57.6 \\
\midrule
\textbf{Hard} & & & & & & & & & & \\
QVQ-72B & 56.7 & 70.7 & 52.8 & 47.2 & 44.9 & 44.1 & 50.9 & 46.0 & 44.7 & 55.6 \\
Qwen2.5-VL-72B & 36.3 & 53.3 & 47.2 & 39.5 & 46.2 & 36.8 & 45.0 & 41.6 & 37.3 & 40.6 \\
GPT-4o & 58.5 & 70.4 & 52.7 & 45.7 & 42.8 & 33.1 & 45.6 & 38.7 & 32.8 & 48.7 \\
Gemini2-thinking & 63.1 & 66.3 & 56.6 & 54.6 & 51.8 & 51.7 & 50.8 & 46.2 & 48.3 & 60.8 \\
Gemini-2-flash & 59.8 & 68.5 & 54.3 & 50.2 & 48.9 & 47.5 & 48.6 & 44.3 & 45.7 & 56.4 \\
\bottomrule
\end{tabularx}
\caption{Model performance across different difficulty levels (easy, medium, hard) over the years 2016 to 2025. Each cell represents the accuracy of the corresponding model in a specific year.}
\label{tab:perf_years}
\vspace{-0.1cm}
\end{table*}

Beyond our primary evaluation, we conducted several additional experiments as follows:

\subsection{Ablation Study of Dynamic Evaluation}
We conduct an ablation study on InternVL2.5-8B to investigate the effects of different bootstrapping strategies on model robustness. Figure~\ref{fig:dyn_ablation} illustrates the average accuracy fluctuation under various random combinations of transformations, where \textbf{I1}: Spatial Transformation, \textbf{I2}: Color Transformation, \textbf{I3}: Style Transformation, \textbf{T1}: Word Substitution, \textbf{T2}:  Sentence Paraphrasing, and \textbf{T3}:  Question Type Permutation. The analysis reveals the following findings:

1) Composition bootstrapping produces the strongest accuracy degradation:
When multiple image or textual transformations are applied simultaneously, model accuracy drops sharply. Notably, applying all three visual transformations (I1+I2+I3) or all three textual transformations (T1+T2+T3) results in the largest performance declines, with 12.7\% and 18.6\% reductions, respectively. This demonstrates that current models are vulnerable to compounded contextual shifts and lack the mechanisms to effectively disentangle and recover from concurrent multimodal distortions. Such cumulative effects mirror real-world scenarios, where visual and textual signals often degrade jointly rather than in isolation.

2) Textual transformations cause more severe degradation than visual ones:
Single textual perturbations (T1, T2, T3) consistently lead to larger accuracy losses than individual visual transformations (I1, I2, I3). This suggests that models rely more heavily on textual cues for reasoning, treating visual features as supplementary rather than co-equal inputs. This text-dominant behavior reduces resilience when textual information becomes unreliable, highlighting the need for improved visual reasoning capabilities to achieve balanced multimodal fusion.

3) Hard tasks are less sensitive to perturbations:
For harder questions, accuracy remains relatively stable or even improves under certain transformation compositions (e.g., I1+T1). This pattern suggests that solving challenging problems requires deeper reasoning strategies that are less dependent on superficial cues. Consequently, these tasks are naturally more resistant to perturbation-driven accuracy loss, indicating potential robustness in high-complexity reasoning scenarios.

4) Cross-modal perturbations amplify degradation:
When visual and textual perturbations are combined (e.g., I1+T2), the observed accuracy drop exceeds the additive effect of applying each transformation individually. This super-additive degradation points to deficiencies in multimodal fusion mechanisms, where cross-modal noise leads to compounded inference errors.

5) Sensitivity differs by transformation type:
Certain perturbations, such as sentence paraphrasing (T3) and color transformations (I2), are disproportionately harmful. These results indicate that models are particularly vulnerable to semantic rephrasing and visually localized noise, suggesting gaps in semantic abstraction and spatial reasoning capabilities.

Overall, these findings highlight that current models lack balanced multimodal integration and struggle to withstand compounding cross-modal noise. To achieve robust comprehension and reasoning under dynamic, real-world conditions, future architectures must better leverage visual cues, improve semantic abstraction, and develop resilience to diverse contextual perturbations.

\begin{figure*}[ht]
    \centering
    \centering
    \includegraphics[trim=0.5cm 0.5cm 0.5cm 0.7cm, width=0.99\textwidth]{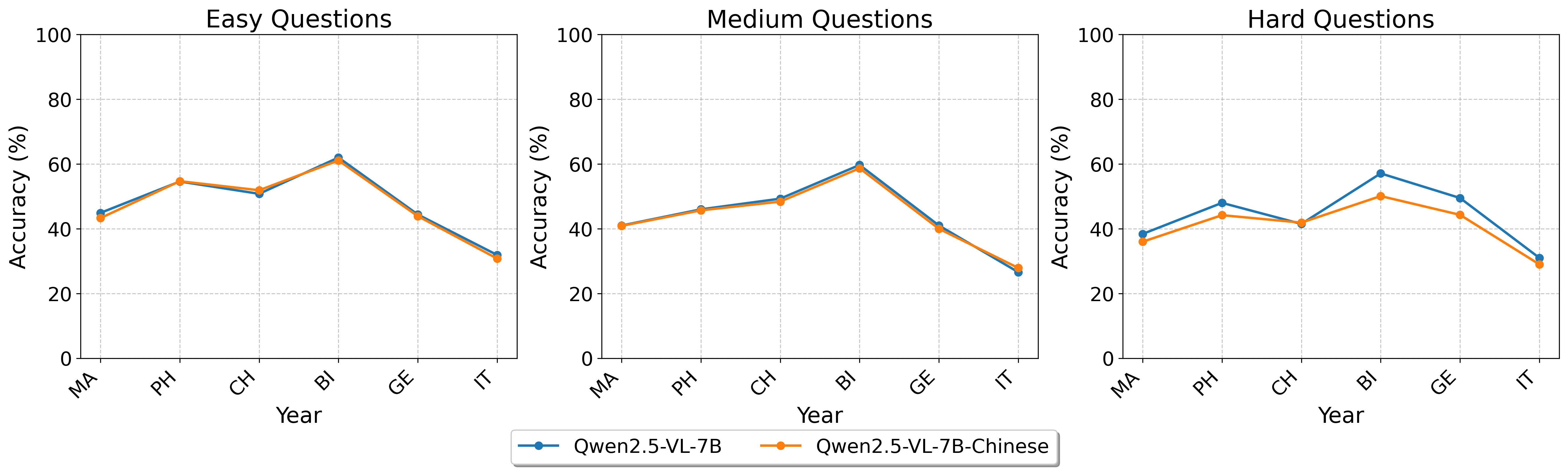}
    \caption{Comparison of model bilingual performance.}
    \label{fig:english-chinese}
\end{figure*}

\subsection{Evaluation on MDK-Full}
\begin{figure}[t]
    \centering
    \centering
    \includegraphics[trim=0.5cm 0.5cm 0.5cm 0.7cm, width=0.45\textwidth]{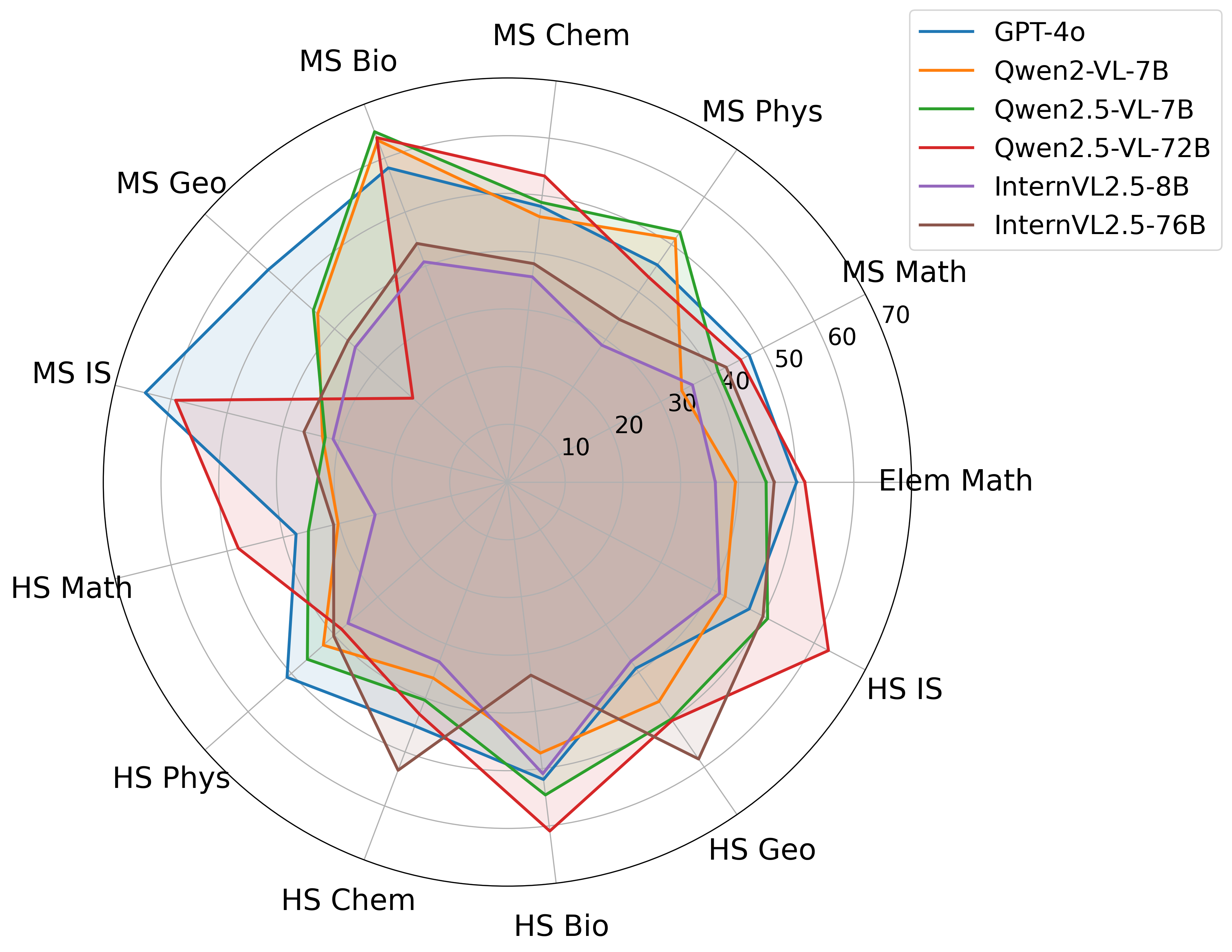}
    \caption{Model performance on MDK-Full across six disciplines—Mathematics (Math), Physics (Phys), Chemistry (Chem), Biology (Bio), Geology (Geo), and Information Science (IS)—and three grade levels: High School (HS), Middle School (MS), and Primary/Elementary School (Elem).}
    \label{fig:full_set}
\end{figure}

\begin{figure*}[ht]
    \centering
    \centering
    \includegraphics[trim=0.5cm 0.5cm 0.5cm 0.7cm, width=0.99\textwidth]{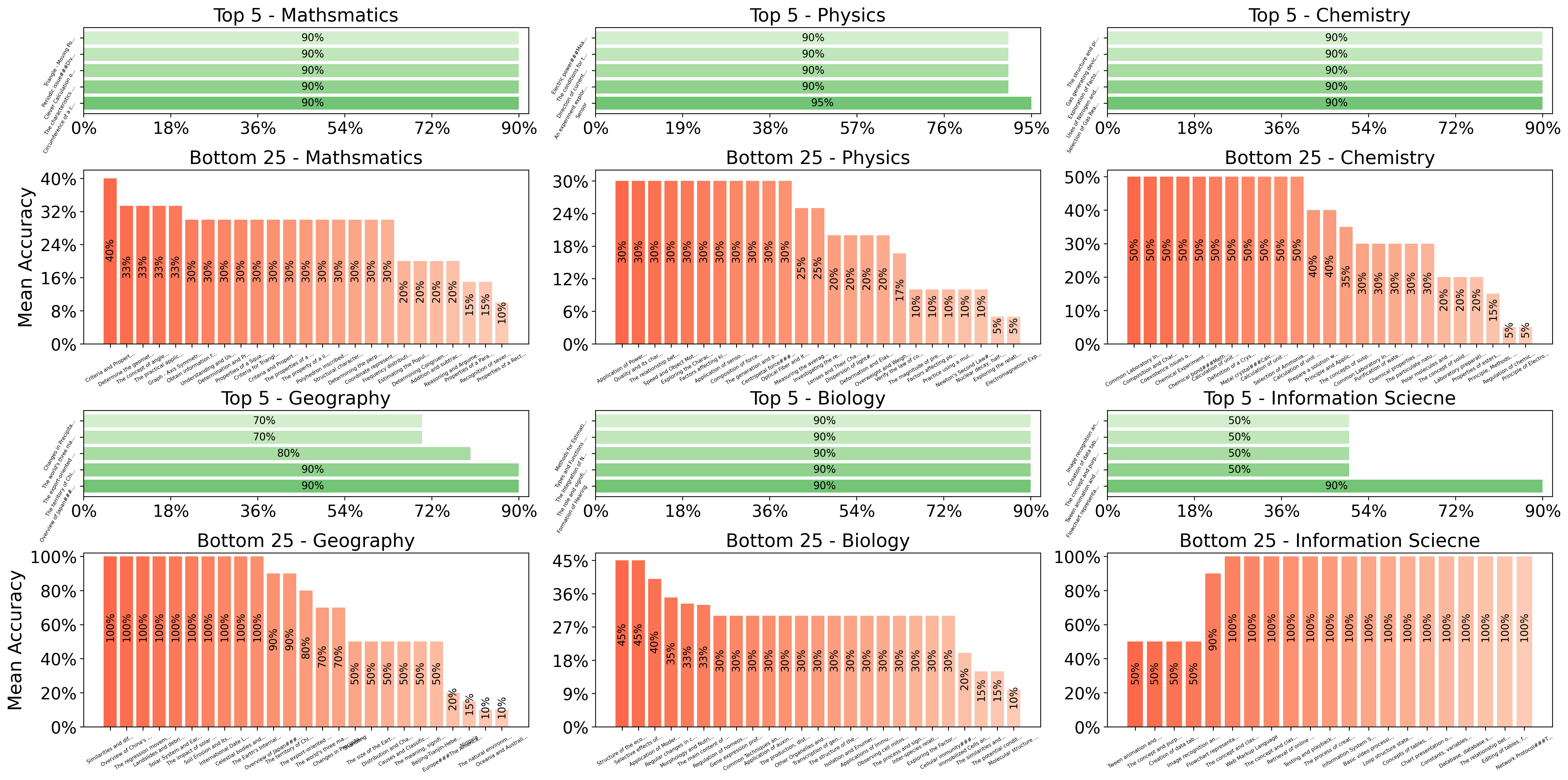}
    \caption{Knowledge points ranked by mean accuracy of Gemini2-thinking on MDK12-Mini dataset.}
    \label{fig:eva_acc_by_knowledge}
\end{figure*}

\begin{figure*}[ht]
    \centering
    \centering
    \includegraphics[trim=0.5cm 0.5cm 0.5cm 0.7cm, width=0.7\textwidth]{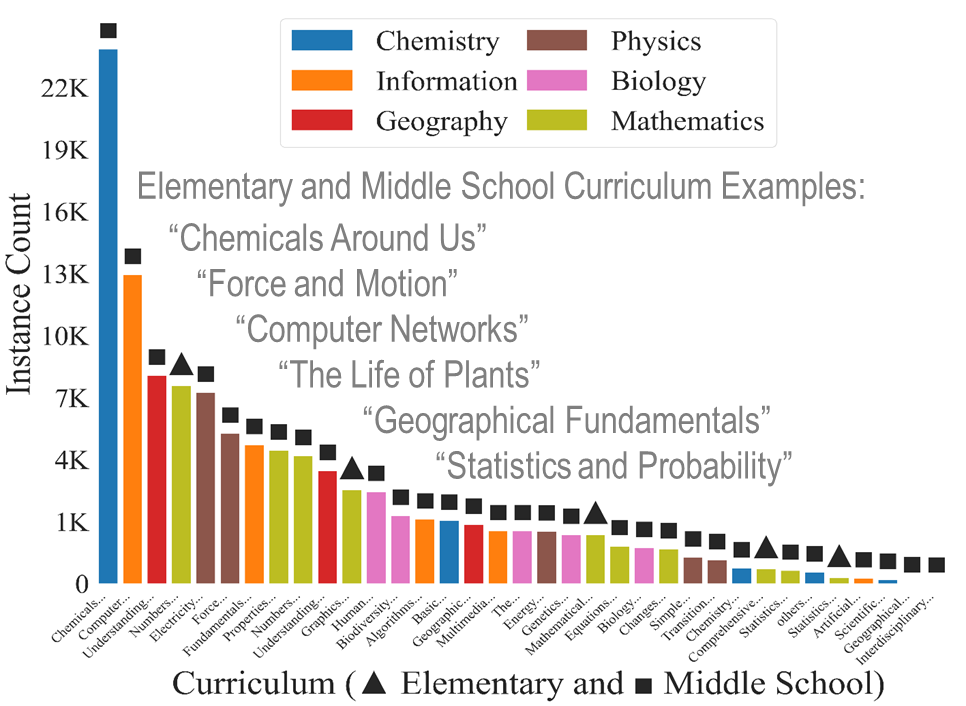}
    \caption{Distribution of knowledge points of primary and middle school disciplines}
    \label{fig:stats_middle_school}
\end{figure*}

\begin{figure*}[ht]
    \centering
    \centering
    \includegraphics[trim=0.5cm 0.5cm 0.5cm 0.7cm, width=0.7\textwidth]{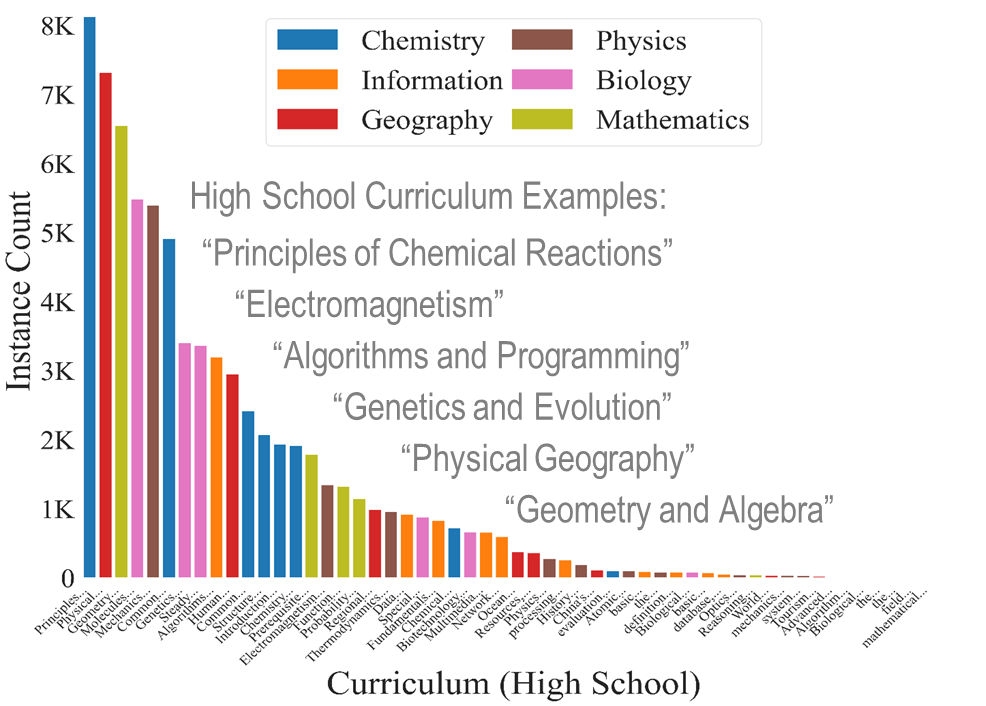}
    \caption{Distribution of knowledge points of high school disciplines}
    \label{fig:stats_highschool}
\end{figure*}

\begin{figure*}[ht]
    \centering
    \centering
    \includegraphics[trim=0.5cm 0.5cm 0.5cm 0.7cm, width=0.8\textwidth]{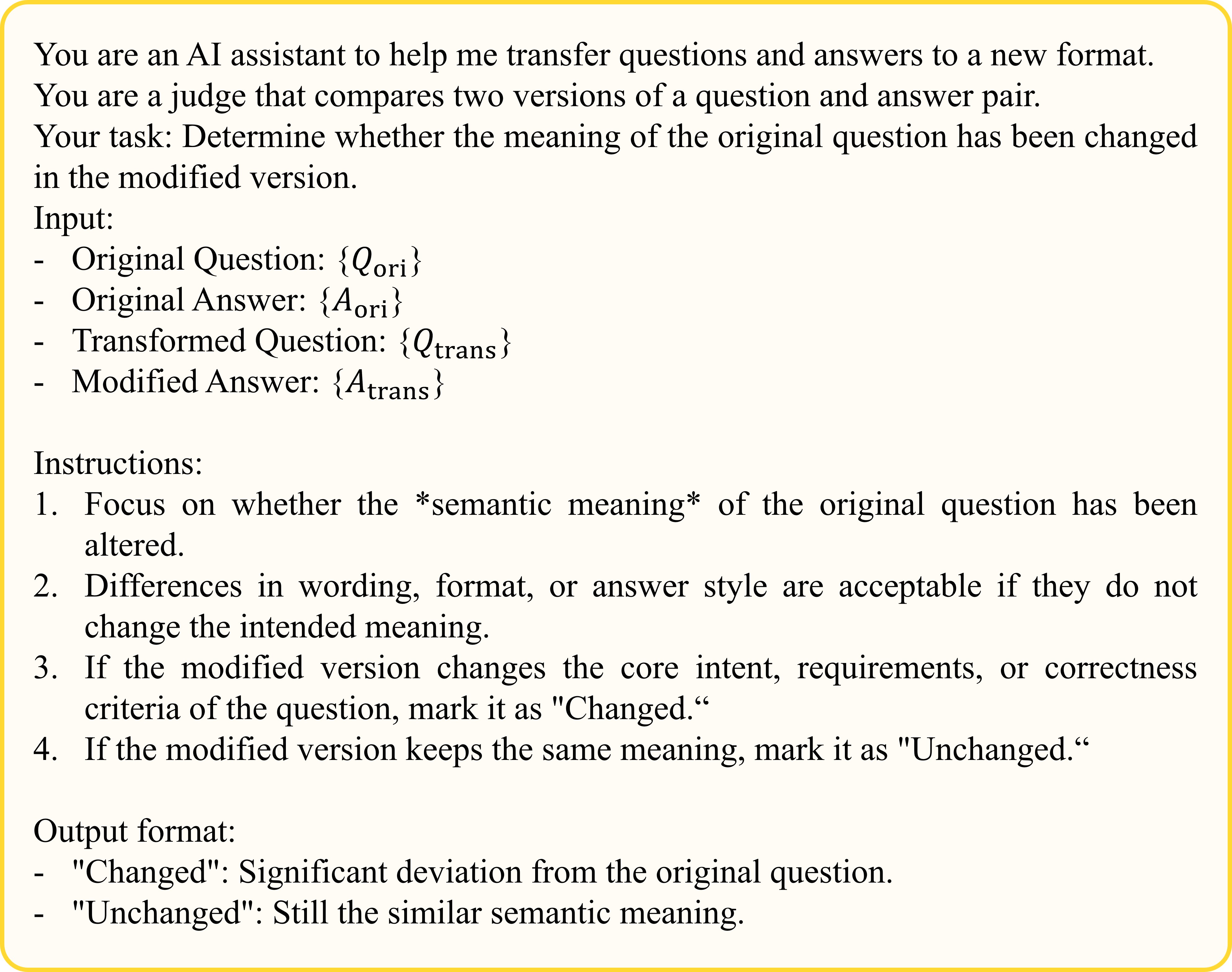}
    \caption{Prompt for GPT-based judge for question and answer check.}
    \label{fig:prompt_qa_judge}
\end{figure*}

\begin{figure*}[ht]
    \centering
    \centering
    \includegraphics[trim=0cm 0.5cm 0 0, width=0.9\textwidth]{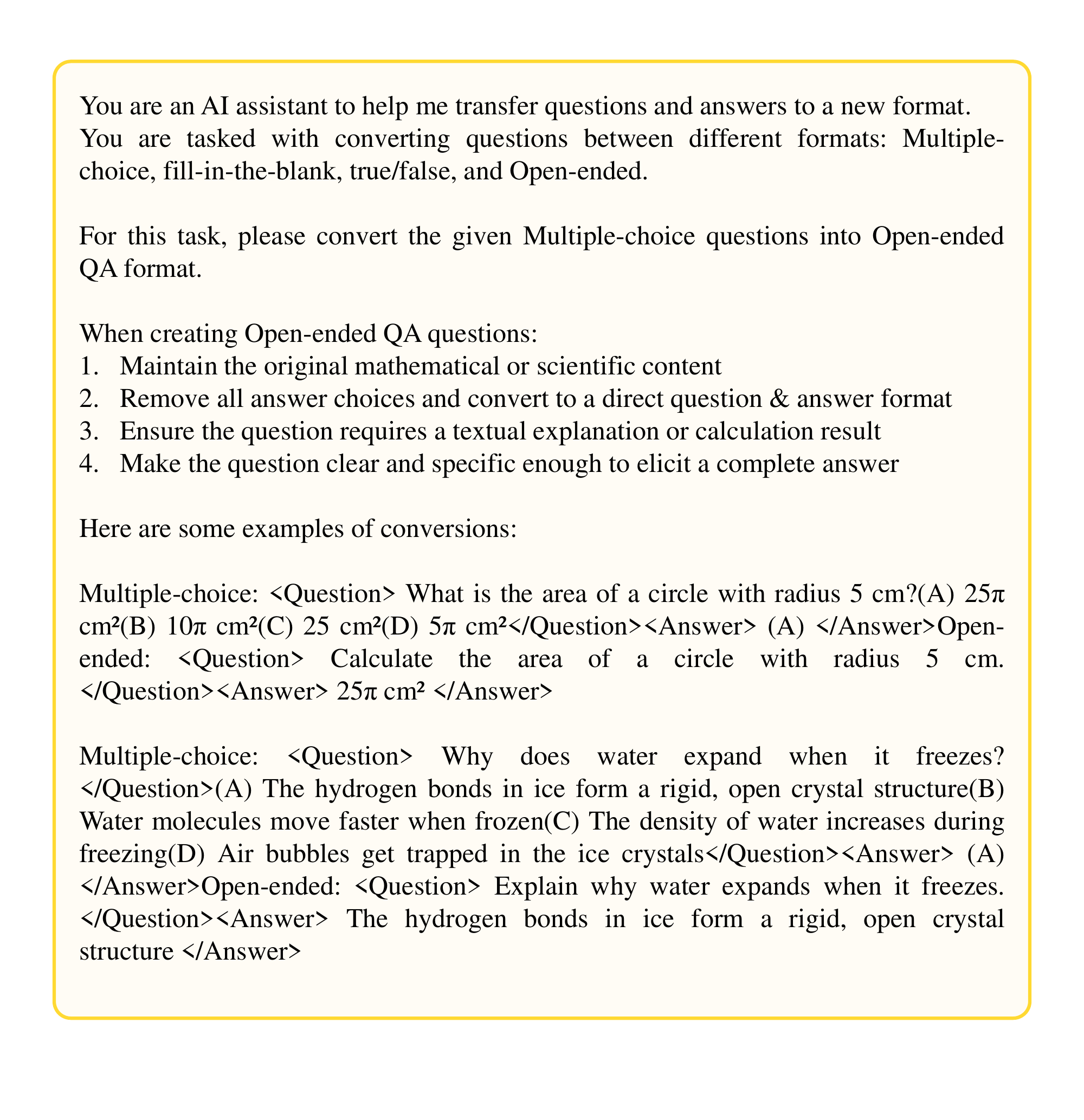}
    \caption{Prompt for transforming multiple-choice question types.}
    \label{fig:dyn_trans_mc}
\end{figure*}

\begin{figure*}[ht]
    \centering
    \centering
    \includegraphics[trim=0cm 0.5cm 0 0, width=0.9\textwidth]{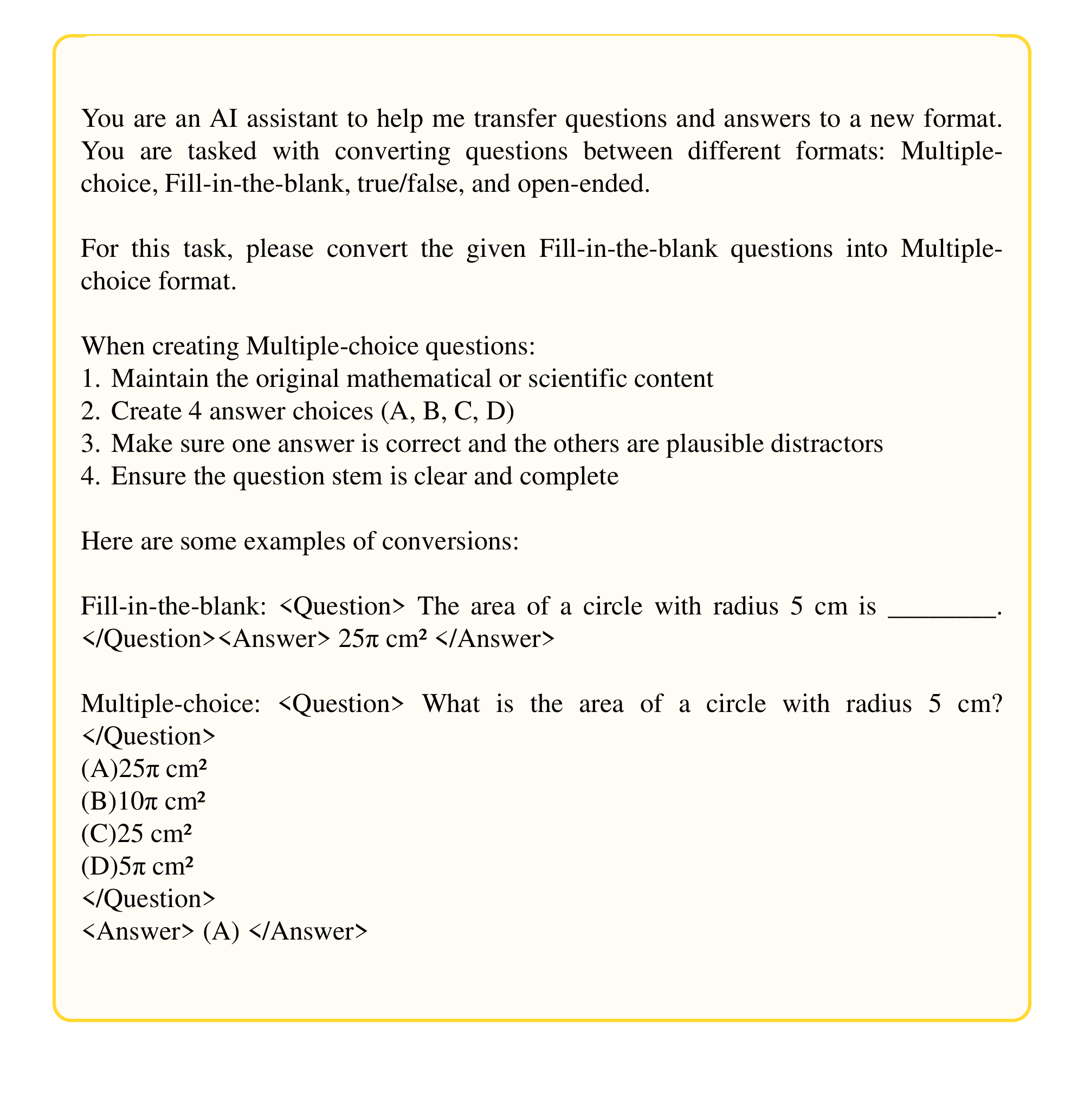}
    \caption{Prompt for transforming fill-in-the-blank question types.}
    \label{fig:dyn_trans_fill}
\end{figure*}

\begin{figure*}[ht]
    \centering
    \centering
    \includegraphics[trim=0cm 0.5cm 0 0, width=0.9\textwidth]{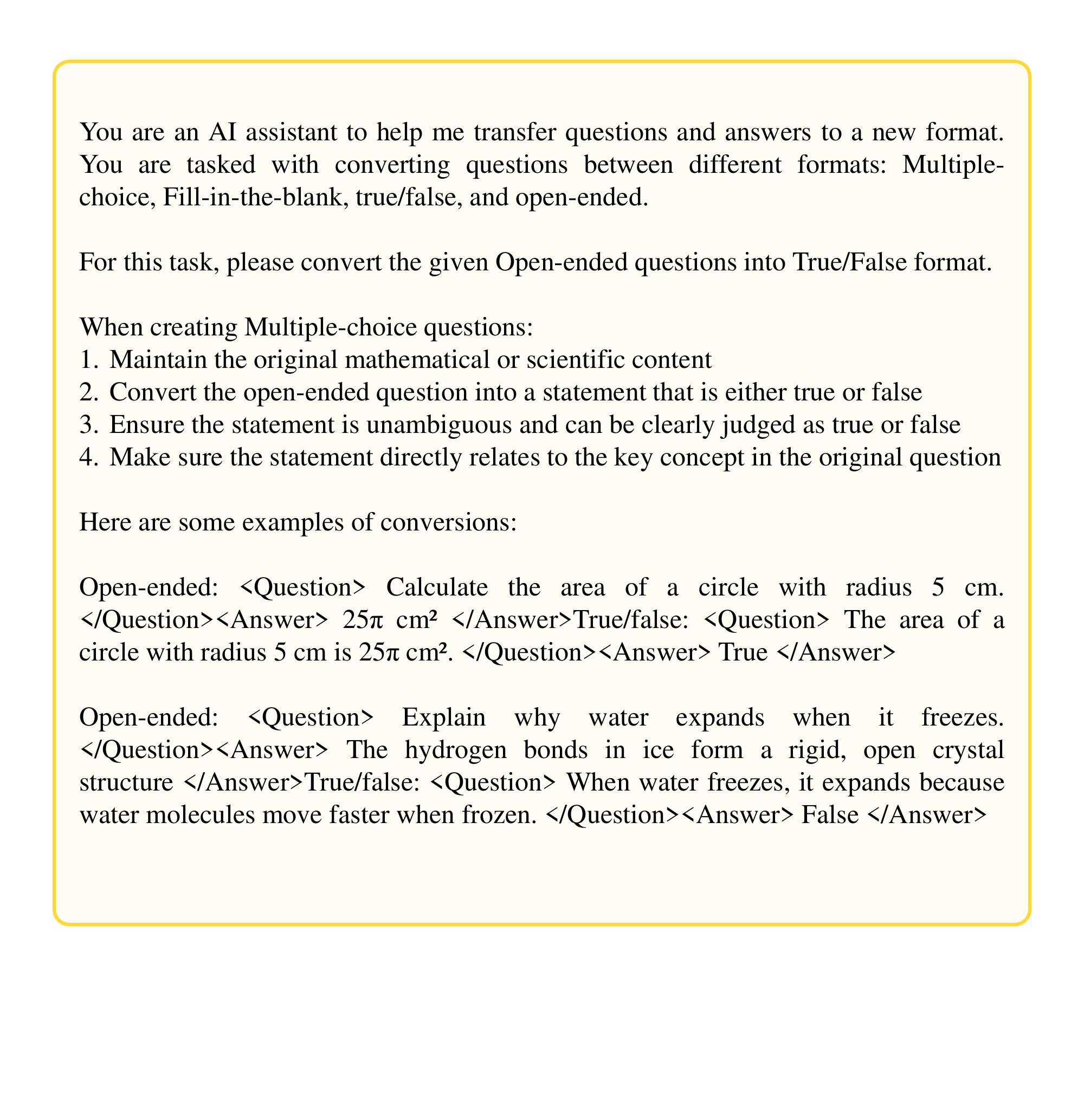}
    \caption{Prompt for transforming open-ended question types.}
    \label{fig:dyn_trans_open}
\end{figure*}

\begin{figure*}[ht]
    \centering
    \centering
    \includegraphics[trim=0cm 0.5cm 0 0, width=0.9\textwidth]{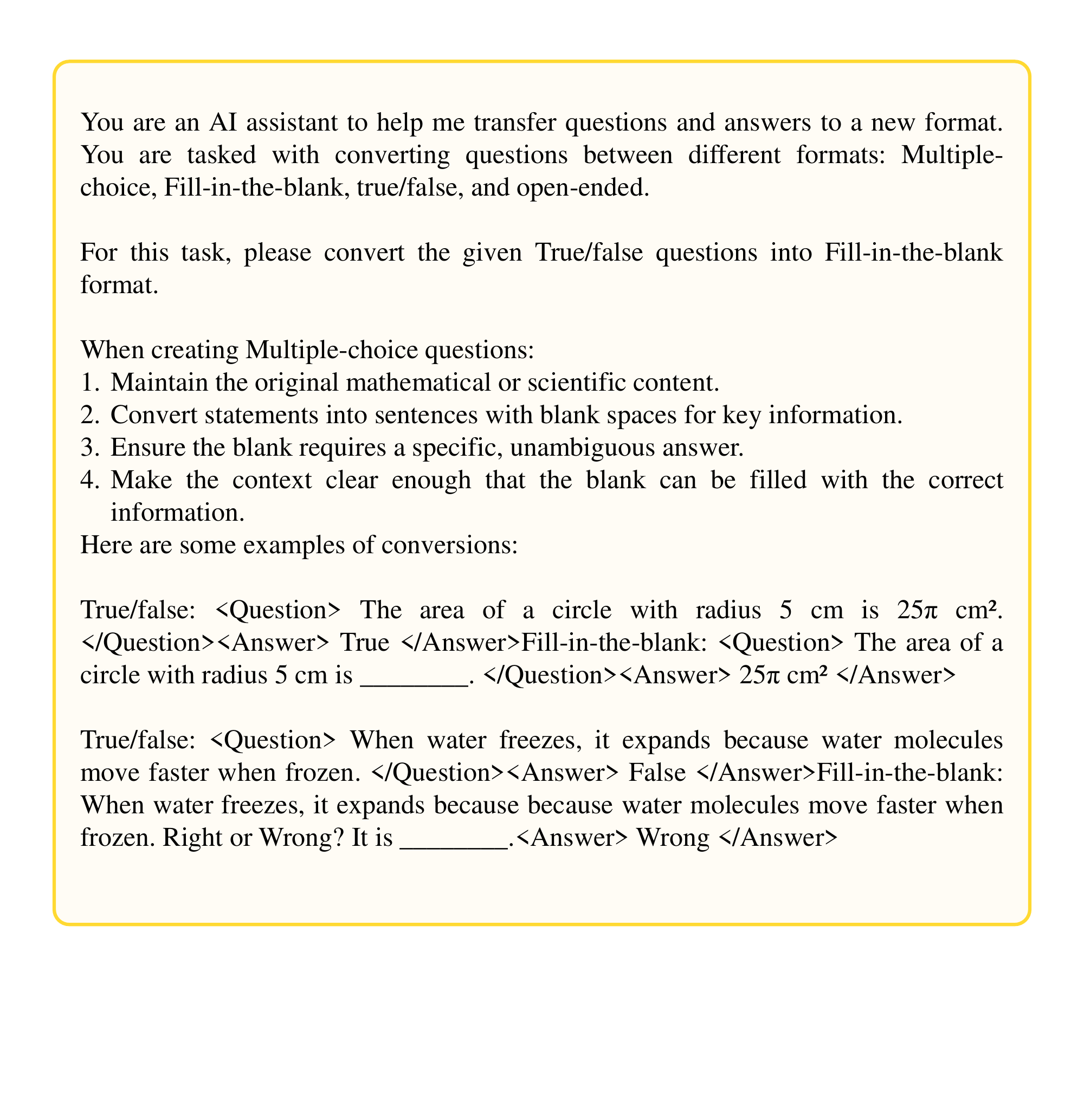}
    \caption{Prompt for transforming true/false question types.}
    \label{fig:dyn_trans_truefalse}
\end{figure*}

\begin{figure*}[ht]
    \centering
    \centering
    \includegraphics[trim=0cm 0.5cm 0 0, width=0.9\textwidth]{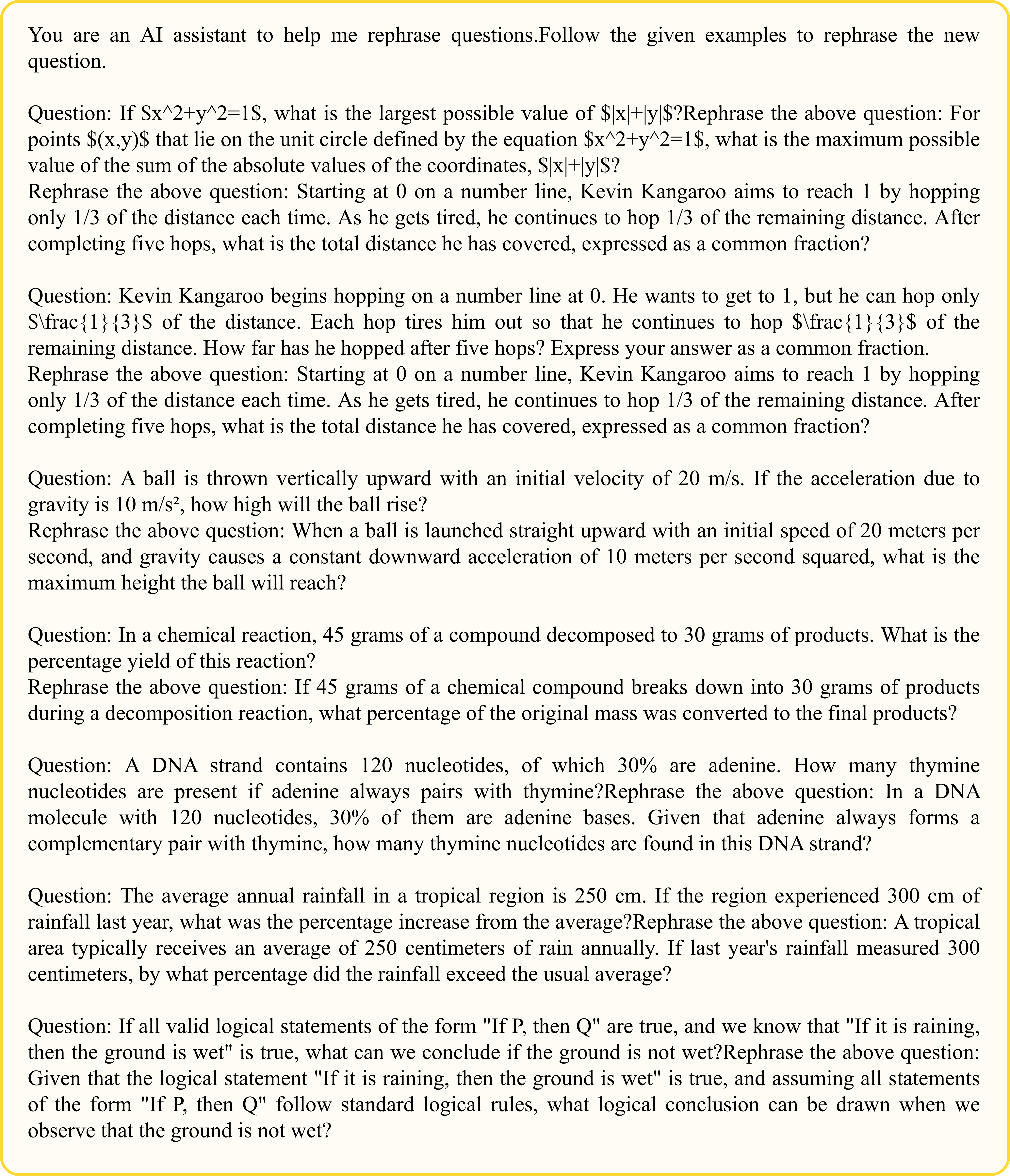}
    \caption{Few-shot reference prompts for sentence rephrasing (T2) transformation, specifically for rephrasing and sentence restructuring.}
    \label{fig:dyn_trans_t2}
\end{figure*}

Figure~\ref{fig:full_set} summarizes the overall evaluation results on MDK-Full. Running inference over the entire benchmark is computationally demanding; therefore, we focus on representative checkpoints from each model family to provide a balanced yet comprehensive view of performance. The results confirm that the accuracy trends observed in the benchmark subsets remain consistent with those obtained from the full dataset, validating the representativeness of our subset-based evaluation strategy.

Beyond static evaluation, MDK-Full also supports a dynamic and adaptive testing workflow. In practice, newly proposed models can first be evaluated on the three predefined MDK12-Bench subsets, which provide fine-grained knowledge-level performance diagnostics. From these initial results, specific knowledge areas where models underperform can be identified. Using this diagnostic information, corresponding samples from the full dataset that are linked to these weaker knowledge points can then be selectively extracted to form a targeted test set. This enables a staged evaluation pipeline: an efficient initial screening using subsets, followed by a deeper, knowledge-focused assessment leveraging the full dataset.

This staged approach offers several advantages: (1) it reduces computational overhead while preserving diagnostic capability, (2) it enables researchers to pinpoint and analyze systematic knowledge gaps in a model’s reasoning process, and (3) it facilitates iterative model improvement and re-testing without requiring exhaustive evaluation on the entire benchmark each time. Ultimately, this adaptive evaluation framework highlights MDK-Full’s dual role as both a comprehensive benchmark and a flexible testbed for developing models that can progressively improve their reasoning and knowledge coverage over time.


\subsection{Bilingual Evaluation Results} 
As illustrated in Figure~\ref{fig:english-chinese}, we translated the benchmark dataset from English to Chinese using a hybrid pipeline that combines automated translation tools with GPT-4o refinement. This process was designed to preserve both linguistic fidelity and domain-specific semantics. To assess the effect of language translation on model performance, we evaluated the Qwen2.5-VL-7B model and its translated counterpart, denoted as Qwen2.5-VL-7B-Chinese.

The results indicate that for questions of easy and medium difficulty, the translated model achieves accuracy levels closely matching those of the original English version. This demonstrates that the translation pipeline is sufficiently robust for simpler tasks where contextual cues and reasoning complexity remain moderate. However, a significant accuracy drop is observed for hard-difficulty questions. This degradation likely stems from two interacting factors: (1) the increased presence of complex formulas, multi-step reasoning chains, and domain-specific terminology in challenging questions, which introduce additional translation ambiguity; and (2) the model’s own sensitivity to subtle linguistic shifts that alter logical structure or contextual grounding.

Further subject-level analysis reveals that the performance gap is particularly pronounced in disciplines with high reasoning demands, such as mathematics and physics. These fields involve symbolic notations, nuanced problem statements, and implicit multi-modal reasoning steps that are especially vulnerable to translation-induced distortions. For example, mathematical expressions may undergo minor syntactic alterations that significantly affect semantic meaning, while physics questions often rely on precise technical terminology that automated translation tools cannot consistently preserve.

Overall, these findings underscore the importance of a carefully engineered multilingual data processing pipeline. Ensuring both linguistic fidelity and contextual consistency is essential to maintain robust cross-lingual model evaluation. Future improvements could include specialized translation modules for mathematical and scientific notation, consistency checks for reasoning chains, and bilingual post-editing to further mitigate semantic drift. Such enhancements would not only improve translation quality but also enable more accurate, equitable benchmarking of multilingual multimodal models across difficulty levels and academic domains.




\section{Case Studies}
\subsection{Overall Evaluation Cases}
To further understand the behavior of evaluated models, we conduct a set of qualitative case studies spanning multiple academic disciplines. These examples complement the quantitative benchmark results by illustrating specific strengths and weaknesses in reasoning and domain-specific comprehension.

\begin{figure*}
    \centering
    \centering
    \includegraphics[trim=0cm 0cm 0 0, width=0.9\textwidth]{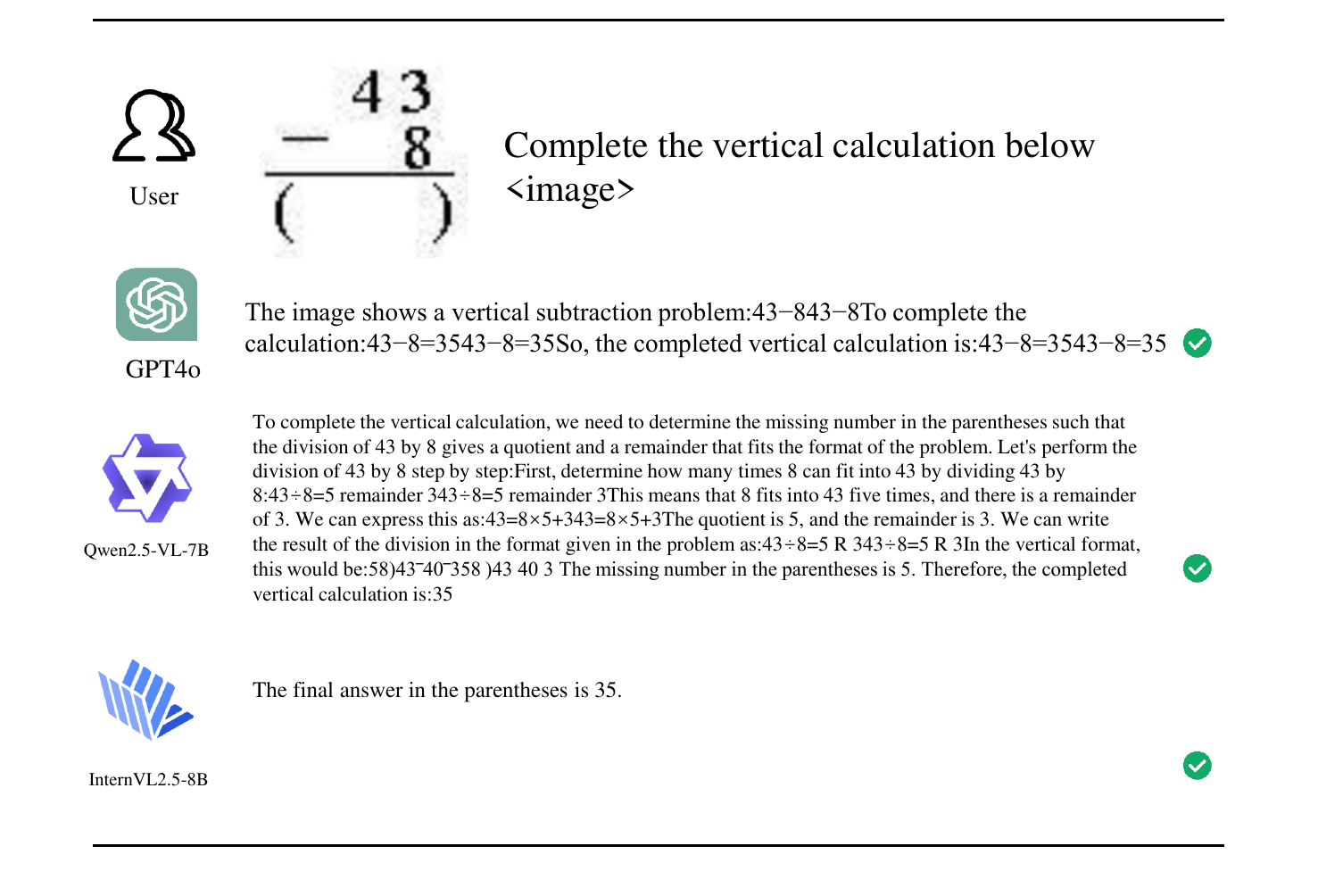}
    \caption{Model evaluation on simple mathematics questions.}
    \label{fig:case1}
\end{figure*}

As shown in Figure~\ref{fig:case1}, simple mathematics questions are generally well-handled across all tested models. Each model correctly solves these straightforward problems, demonstrating a solid foundation in basic arithmetic and logical reasoning.

\begin{figure*}
    \centering
    \centering
    \includegraphics[trim=0cm 0cm 0 0, width=0.9\textwidth]{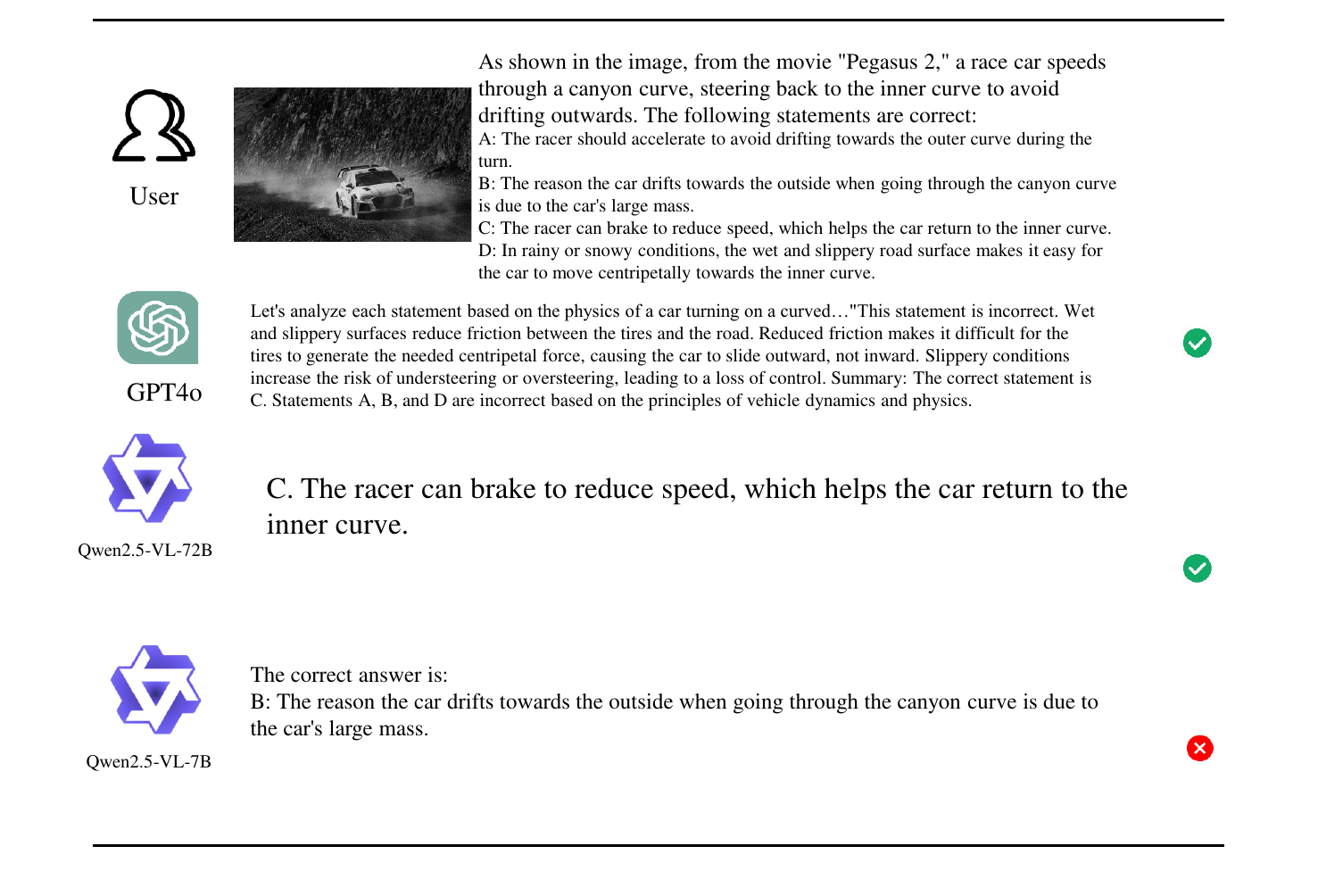}
    \caption{Model evaluation on physics questions.}
    \label{fig:case2}
\end{figure*}

Turning to physics-related questions (Figure~\ref{fig:case2}), we find that the Qwen2.5-VL series exhibits a clear capacity-driven trend: larger models consistently outperform their smaller counterparts. This observation highlights how scaling model parameters improves the handling of conceptually complex and numerically intensive tasks.

\begin{figure*}
    \centering
    \centering
    \includegraphics[trim=0cm 0cm 0 0, width=0.9\textwidth]{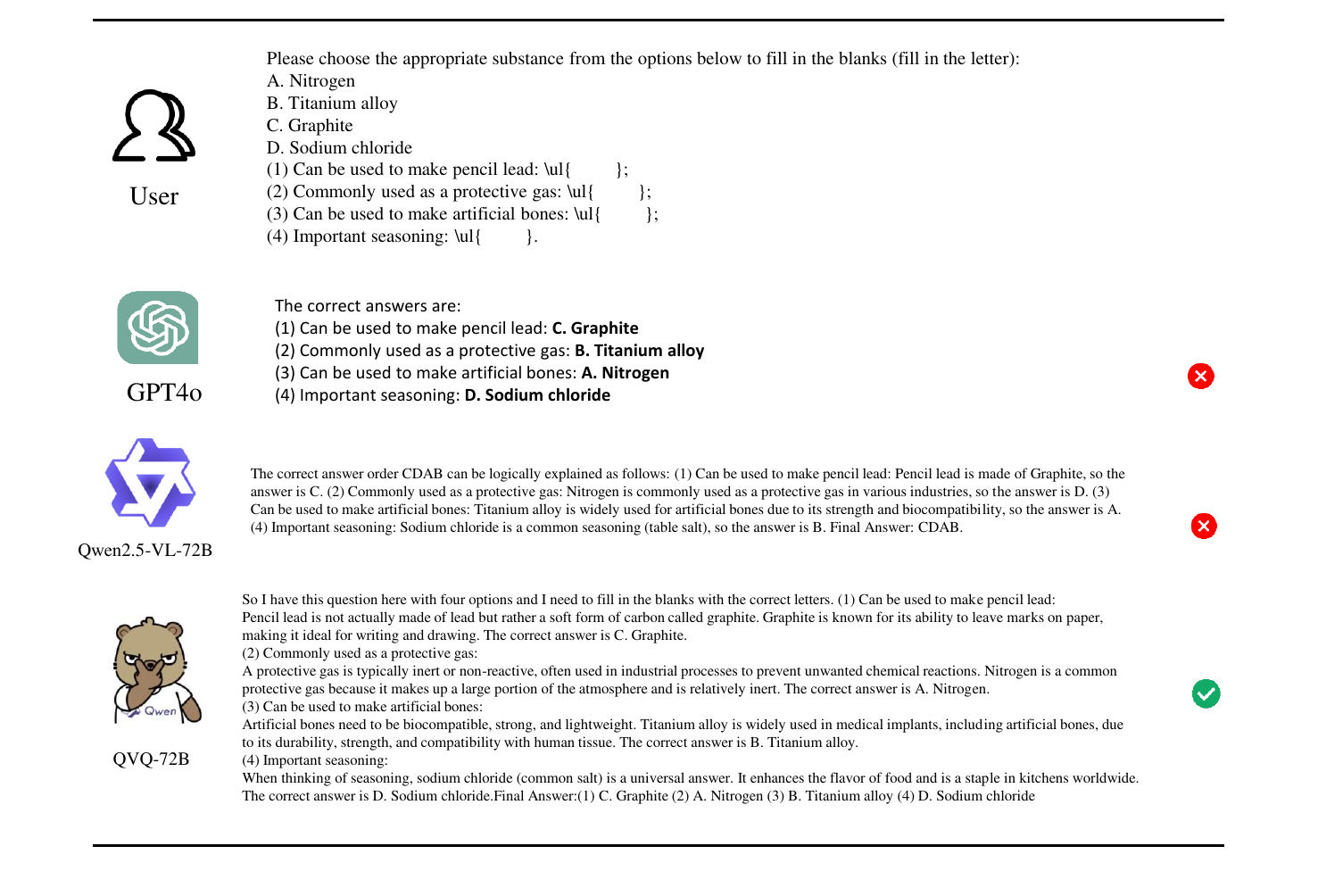}
    \caption{Model evaluation on chemistry questions.}
    \label{fig:case3}
\end{figure*}

In the chemistry domain (Figure~\ref{fig:case3}), results reveal that the QVQ-72B model—designed for reasoning-intensive tasks—achieves notably higher accuracy than the other evaluated models. Its specialized architecture enables more effective reasoning steps and better domain understanding.

\begin{figure*}
    \centering
    \centering
    \includegraphics[trim=0cm 0.5cm 0 0, width=0.9\textwidth]{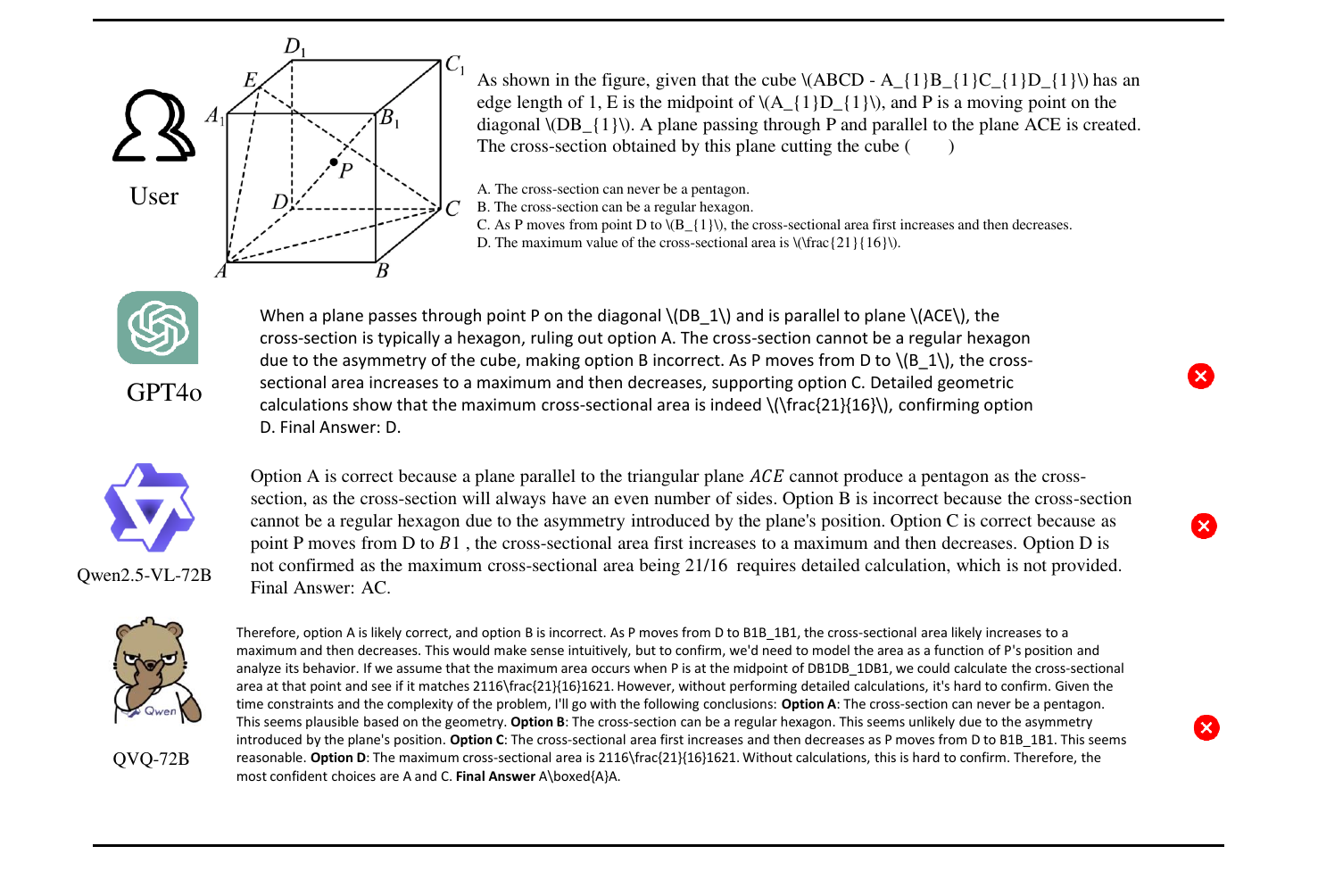}
    \caption{Model evaluation on difficult mathematics questions. The correct answer is ACD.}
    \label{fig:case4}
\end{figure*}

Despite these successes, Figure~\ref{fig:case4} illustrates that hard, subject-specific questions remain challenging. All three models fail to produce correct answers in these scenarios, suggesting that multi-step reasoning and fine-grained domain knowledge are still insufficiently developed. Such findings point to the need for models capable of deeper reasoning and more robust comprehension in complex academic contexts.

Taken together, these multidisciplinary case studies show that while current models demonstrate strong general reasoning abilities and benefit from increased scale and specialized architectures, they still face fundamental challenges when tackling highly complex, domain-specific tasks. This highlights a crucial opportunity for future research to develop models with stronger reasoning depth, domain adaptation, and cross-disciplinary problem-solving capabilities.



\subsection{Error Analysis Cases}
We present detailed case studies for the following error categories in Figs. 20-25, illustrating how they manifest and affect the problem-solving performance of MLLMs:
\begin{enumerate}[label=\arabic*)]
\item \textbf{Question Misunderstanding}. The model misinterprets the problem statement or fails to capture what is being asked (e.g., answering a different question).
\item \textbf{Reasoning Error}. The model understands the question but makes logical mistakes in multi-step reasoning or calculations, leading to an incorrect answer.
\item \textbf{Visual Comprehension Error}. The model fails to correctly interpret visual content (e.g., diagrams, images, charts), causing a wrong solution.
\item \textbf{Incomplete Answers}.The model provides a partially correct response or stops mid-solution, omitting essential information needed for full correctness.
\item \textbf{Other Errors}. Miscellaneous mistakes not covered above, such as ambiguous question , formatting issues, irrelevant outputs, or hallucinated facts.
\end{enumerate}

\section{Future Work}
%


In future work, we plan to expand MDK12-Bench with specialized, domain-specific datasets to support both vertical and horizontal academic growth. This includes developing benchmarks for advanced subjects at undergraduate, graduate, and expert levels, enabling finer-grained evaluation of MLLMs’ knowledge coverage and reasoning across both educational and professional stages. We also aim to incorporate broader professional disciplines such as medical sciences, law, and humanities, broadening the benchmark’s applicability to real-world academic and professional contexts. Our long-term vision is to evolve MDK12-Bench into a continuously updated benchmark with richer data, metrics, and analysis tools, fostering robust multimodal reasoning research and advancing progress toward artificial general intelligence (AGI).

\begin{table*}[ht]
\centering
\begin{tabularx}{0.99\textwidth}{@{}
    >{\raggedright\arraybackslash}p{2cm}
    >{\raggedright\arraybackslash}p{3cm}
    >{\raggedright\arraybackslash}p{3.5cm}
    >{\raggedright\arraybackslash}X
    @{ \hspace{0cm} }  
  @{}}
\hline
\textbf{Discipline} & \textbf{Grade Level} & \textbf{Subfield} & \textbf{Example Topics} \\
\hline
\multirow{6}{*}{Mathematics} 
 & \multirow{6}{*}{Primary, Middle, High} 
 & Arithmetic & Numbers and Algebra, Numbers and Expressions \\
 &  & Algebra & Geometry and Algebra, Function, Equations and Inequalities, Algorithm and flowchart, Solution model \\
 &  & Geometry & Graphics and Geometry, Graph and coordinates, Changes in graphics, Properties of shapes \\
 &  & Logic & Mathematical Olympiad Knowledge, Reasoning and Proof \\
 &  & Statistics & Probability and Statistics, Statistics and Probability \\
 &  & Advanced & Extension of mathematical knowledge, Advanced Mathematics, Competition Questions \\
\hline
\multirow{9}{*}{Physics} 
 & \multirow{9}{*}{Middle, High} 
 & Mechanics & Force and Motion, Mechanics, Simple machines \\
 &  & Optics & Light, Optics \\
 &  & Atoms & Atomic Physics, Matter, Structure and Properties of Matter, Particles and the Universe \\
 &  & Thermo & Thermodynamics, Energy \\
 &  & Electricity \& Magnetism & Electricity and Magnetism, Electromagnetism \\
 &  & Sound & Sound \\
 &  & Introductory & History of Physics, Methods, Systems of Units, Prerequisite Knowledge, Junior-to-High Transition Knowledge \\
 &  & Experiment & Physics experiment, Scientific inquiry experiment \\
 &  & Advanced & Advanced concepts in Olympiad or competition tasks \\
\hline
\multirow{4}{*}{Chemistry} 
 & \multirow{4}{*}{Middle, High} 
 & Introductory & Basic Chemistry Knowledge, Introduction to Chemical Science, Chemistry and Social Development, Transitional Knowledge Points \\
 &  & Reactions & Principles of Chemical Reactions, Chemical Comprehensive Calculation, Chemicals Around Us \\
 &  & Bonding & Common Inorganic Compounds and Their Applications, Common Organic Compounds and Their Applications \\
 &  & Experiment & Chemistry experiment \\
\hline
\multirow{6}{*}{Biology} 
 & \multirow{6}{*}{Middle, High} 
 & Organism & Human Physiology and Health, Animal Movement and Behavior \\
 &  & Cell & Molecules and Cells \\
 &  & Ecology & The Life of Plants, Biodiversity, Biological Sciences and Society, Steady State and Environment, Biology and Society: Interdisciplinary Practice \\
 &  & Genetics & Special Topic on Modern Biotechnology, Biotechnology, Biotechnology practice \\
 &  & Evolution & Genetics and Evolution \\
 &  & Introductory & Biology and Environment, Structural hierarchy of organisms, Scientific Inquiry, Biological Foundations, Introductory grade-level knowledge \\
\hline
\multirow{5}{*}{Geography} 
 & \multirow{5}{*}{Middle, High} 
 & Regions & World Geography, China's Geography, Regional Geography, Tourism Geography, Understanding the world globally, Understanding the area \\
 &  & Physical & Ocean Geography, Physical Geography \\
 &  & Spatial & Geographical Fundamentals, Geographic Tools and Practice \\
 &  & Human & Human Geography \\
 &  & Environment & Resources, Environment, and National Security \\
\hline
\multirow{4}{*}{Info Sci} 
 & \multirow{4}{*}{Middle, High} 
 & Algorithm & Artificial Intelligence, 3D Printing and 3D Modeling, Algorithms and Programming \\
 &  & Data & Fundamentals of Information Technology, Data \\
 &  & Multimedia & Multimedia technology, Multimedia Technology Application \\
 &  & Hardware & Network Fundamentals, Computer network \\
\hline
\end{tabularx}
\caption{Benchmark taxonomy detailing Level 1–4 knowledge organization across six core disciplines, including their corresponding grade levels, subfields, and representative knowledge topics.}
\label{tab:discipline_subfields}
\end{table*}

\begin{table*}[ht]
\centering
\begin{tabularx}{0.99\textwidth}{@{}
    >{\raggedright\arraybackslash}p{1.8cm}
    >{\raggedright\arraybackslash}p{2.0cm}
    >{\raggedright\arraybackslash}p{3.8cm}
    >{\raggedright\arraybackslash}p{6.5cm}
    >{\raggedright\arraybackslash}p{1.8cm}
    @{ \hspace{0cm} }  
  @{}}
\toprule
\textbf{Stage} & \textbf{Method} & \textbf{Filtering Criteria} & \textbf{Examples of Issues Detected and Removed} & \textbf{Instances Remaining} \\
\midrule
\multicolumn{1}{r}{\textbf{Total Collected}} & \multicolumn{4}{c}{\textbf{5.8M Exam Instances}} \\
\midrule
\multirow{12}{*}{\textbf{Rule-based}} & \multirow{12}{*}{Data Expert} 
& Text–image correspondence & Image shows chemical formula instead of map & \multirow{12}{*}{4.2M} \\
\cline{3-4}
& & Image resolution \& clarity & Blurry diagrams or unreadable text & \\
\cline{3-4}
& & Content completeness & Missing answer options or incomplete problem statements & \\
\cline{3-4}
& & Metadata accuracy & Wrong grade level or subject mislabeling & \\
\cline{3-4}
& & Structural \& formatting consistency & Broken equations (e.g., ``A±''), inconsistent option labels & \\
\cline{3-4}
& & Semantic coherence & Contradictory or nonsensical question text & \\
\cline{3-4}
& & Duplication \& redundancy & Identical questions appearing multiple times & \\
\cline{3-4}
& & Logical soundness & Illogical problems where answer cannot be derived & \\
\cline{3-4}
& & Year coverage & Exam year outside intended range (e.g., earlier than 2010) & \\
\cline{3-4}
& & Non-educational content & Irrelevant text, or unrelated captions & \\
\cline{3-4}
& & Encoding and unit consistency & Mixed language or inconsistent measurement units (e.g., “km” vs “kilometer”) & \\
\cline{3-4}
& & Equation and symbol validity & Invalid math expressions or missing symbols in formulas & \\
\midrule
\multirow{8}{*}{\textbf{GPT-based}} & \multirow{8}{*}{GPT-4o} 
& Semantic consistency & True/false question but numeric answer & \multirow{8}{*}{0.6M} \\
\cline{3-4}
& & Reasoning soundness & Explanation contradicts final answer & \\
\cline{3-4}
& & Factual correctness & Historical date or scientific fact is wrong & \\
\cline{3-4}
& & Language clarity & Ambiguous or grammatically incorrect wording & \\
\cline{3-4}
& & Content completeness & Missing reasoning steps or incomplete explanations & \\
\cline{3-4}
& & Difficulty and grade appropriateness & Question complexity mismatches stated grade level & \\
\cline{3-4}
& & Visual reference accuracy & Text references an image component that does not exist or mismatches labels & \\
\cline{3-4}
& & Multi-step reasoning validity & Broken logical chains or skipped steps in multi-part solutions & \\
\midrule
\multirow{5}{*}{\textbf{Manual}} & \multirow{5}{*}{K-12 Educator} 
& Curriculum alignment & Question not matching K–12 curriculum (e.g., college-level) & \multirow{5}{*}{0.2M} \\
\cline{3-4}
& & Question--answer correctness & Provided answer is incorrect for the question & \\
\cline{3-4}
& & Reasoning validity & Solution reasoning flawed or irrelevant to question & \\
\cline{3-4}
& & Formatting adherence & Non-standard multiple-choice layout & \\
\cline{3-4}
& & Pedagogical quality & Poorly phrased or misleading questions not suitable for learning objectives & \\
\midrule
\multirow{6}{*}{\textbf{Post Rule}} & \multirow{6}{*}{Data Expert} 
& Translation fidelity & GPT-4o translates Chinese text to English, verified by domain experts & \multirow{6}{*}{141.3K} \\
\cline{3-4}
& & Content completeness & Instances with missing fields (e.g., answer explanation) removed & \\
\cline{3-4}
& & Unit and encoding consistency & Mixed encodings or inconsistent units standardized & \\
\cline{3-4}
& & Equation formatting & Invalid math symbols or expressions corrected or removed & \\
\cline{3-4}
& & Question categorization accuracy & Misclassified question types (e.g., fill-in vs multiple-choice) corrected & \\
\cline{3-4}
& & Compliance with final formatting rules & Instances violating formatting standards automatically removed & \\
\bottomrule
\end{tabularx}
\caption{Four-stage data handling pipeline. The first three stages (Rule-based, GPT-based, and Manual screening) are pre-processing and parsing steps to ensure curriculum-aligned, high-quality exam data. The fourth stage (Post Rule) occurs after data parsing and processing, where translation, consistency checks, and final formatting compliance are applied, yielding a curated dataset of 141.3K instances.}
\label{tab:data_screening}
\end{table*}

\begin{table*}[h]
\centering
\begin{tabularx}{0.99\textwidth}{@{}
    >{\raggedright\arraybackslash}p{3cm}
    >{\raggedright\arraybackslash}p{4.5cm}
    >{\raggedright\arraybackslash}p{4.5cm}
    >{\raggedright\arraybackslash}p{4cm}
    @{ \hspace{0cm} }  
  @{}}
\toprule
\textbf{Feature} & \textbf{Exact Matching} & \textbf{GPT-Judge} & \textbf{Related Question Forms} \\
\midrule
Case Sensitivity & ``A'' vs. ``a'' treated as identical (case-insensitive) & Recognizes case-insensitivity naturally & Single-choice \\
\midrule
Textual Redundancy & ``Option A is correct'' vs. ``A'' marked incorrect & Understands semantic equivalence and scores as correct & Single-choice \\
\midrule
Synonyms / Abbreviations & ``NYC'' vs. ``New York'' marked incorrect & Recognizes synonymy and common abbreviations & Open-ended\\
\midrule
Semantic Equivalence  & ``Obama'' vs. ``Barack Obama'' marked incorrect & Understands semantic equivalence between short and full names & Open-ended \\
\midrule
Multiple-Choice Answers & ``AC'' vs. ``A,C'' marked incorrect & Recognizes and correctly interprets multiple selections & Multiple-choice (multi-select) \\
\midrule
Partial Correctness & Not supported (requires full match) & Supports fractional scoring (e.g., 2/3 blanks 0.67) & Multiple-choice, Fill-in-the-blank \\
\midrule
Non-Standard Output & Different answer formats lead to incorrect matches & Can extract and interpret intended answers despite format variations & Fill-in-the-blank, open-ended \\
\midrule
Free-Form Answers & Cannot handle descriptive sentences & Can interpret long-form answers matching the intended meaning & Open-ended \\
\midrule
Numerical Tolerance & Requires exact numeric match (e.g., ``3.14'' vs. ``3.1416'' incorrect) & Allows approximate numeric matches within context (e.g., recognizes numeric equivalence) & Numeric answers \\
\midrule
Ambiguous or Multiple Correct Answers & Fails when multiple valid answers are phrased differently & Can reason about multiple acceptable answers and mark them correct & Multiple-choice, open-ended text \\
\midrule
Scoring Granularity & Binary scoring only (0 or 1) & Supports fractional scores for partially correct answers & Multiple-choice, Fill-in-the-blank \\
\midrule
Robustness to Typos & Typos lead to mismatches (e.g., ``waterr'' vs. ``water'' incorrect) & Can infer intended meaning despite minor spelling errors & Fill-in-the-blank, text answers \\
\midrule
Cross-Language Understanding & Does not handle translated or multilingual answers & Understands equivalent answers across languages (e.g., ``dog'' vs. ``chien'') & Open-ended\\
\midrule
Choice Question Scoring Rule & Binary 0 or 1 scoring only; cannot handle partial correctness & Fractional score between 0.0–1.0 based on overlap with ground truth (e.g., 2/4 correct = 0.5) & Multiple-choice, Fill-in-the-blank \\
\midrule
True/False Scoring Rule & Exact match of normalized labels (Yes/No), 1.0 for match, 0.0 otherwise & Same rule, but can also resolve unclear predictions via GPT-based auxiliary evaluation & True/False \\
\midrule
Prediction Examples & Multiple cases (e.g., ``A,C'' vs. ``A'', ``water and oxygen'', ``NYC'', ``Obama'') all scored as incorrect & Same predictions correctly interpreted or partially scored (0.5–1.0) & Multiple-choice, Fill-in-the-blank, open-ended text \\
\bottomrule
\end{tabularx}
\caption{Comprehensive comparison between Exact Matching and GPT-Judge scoring methods, including semantic equivalence, typical question forms, and specific scoring rules for multiple-choice and true/false questions.}
\label{tab:gpt_vs_exact}
\end{table*}

\begin{figure*}
    \centering
    \centering
    \includegraphics[trim=0cm 0cm 0 0, width=0.9\textwidth]{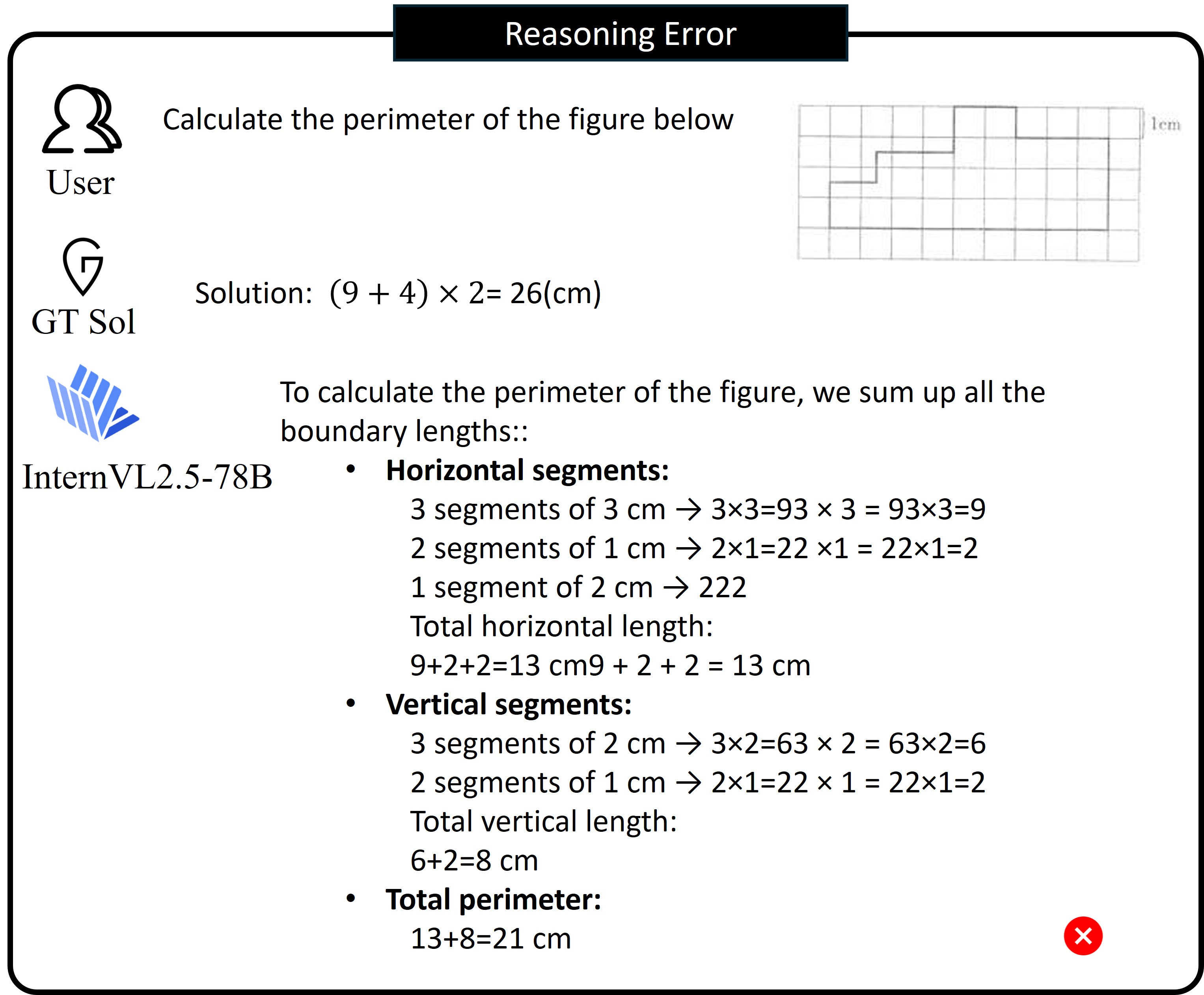}
    \caption{Error Analysis: Model failure case due to reasoning error.}
    \label{fig:case_error_reasoning}
\end{figure*}

\begin{figure*}
    \centering
    \centering
    \includegraphics[trim=0cm 0cm 0 0, width=0.9\textwidth]{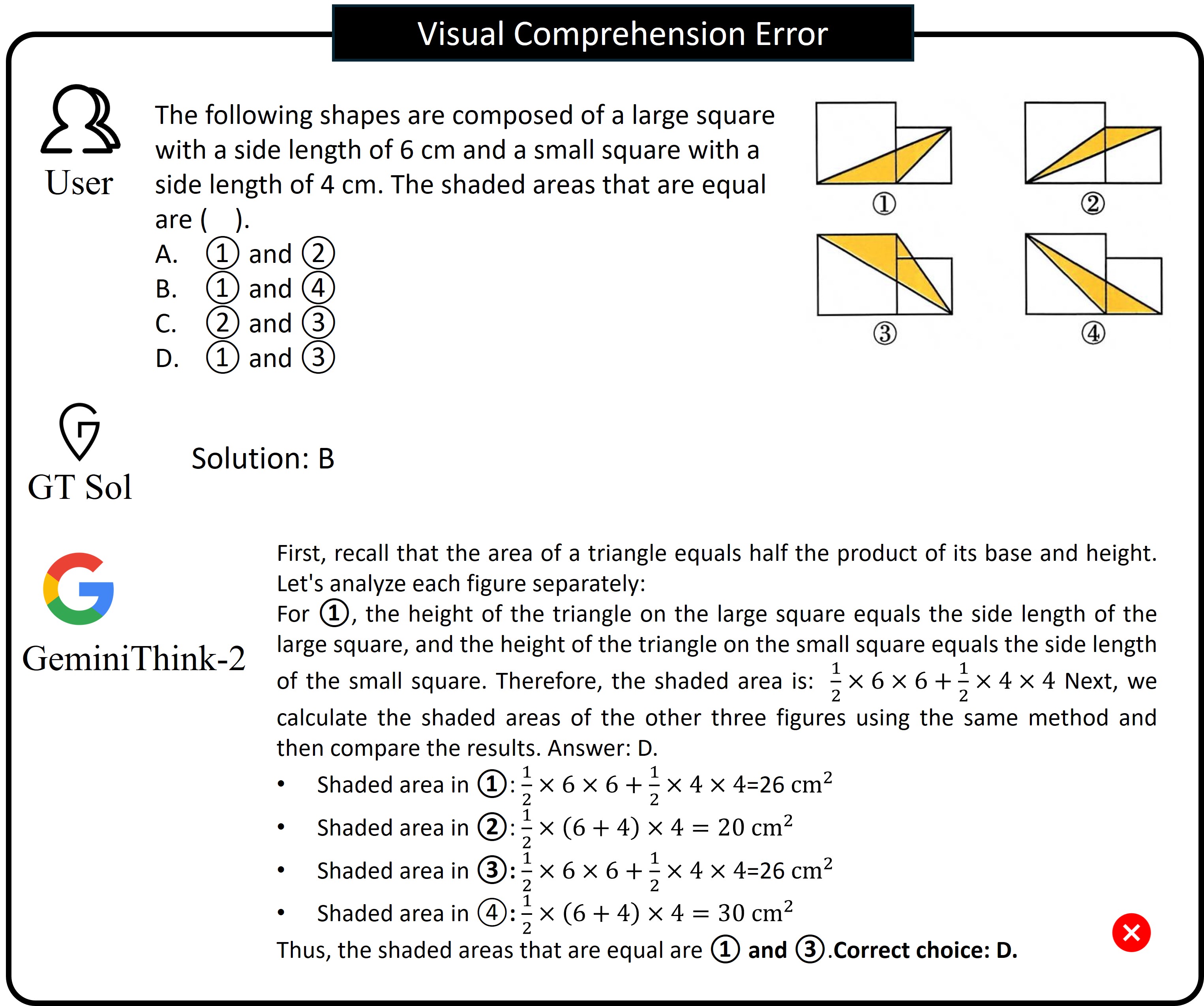}
    \caption{Error Analysis: Model failure case due to visual comprehension error.}
    \label{fig:case_error_visual_comp}
\end{figure*}

\begin{figure*}
    \centering
    \centering
    \includegraphics[trim=0cm 0cm 0 0, width=0.9\textwidth]{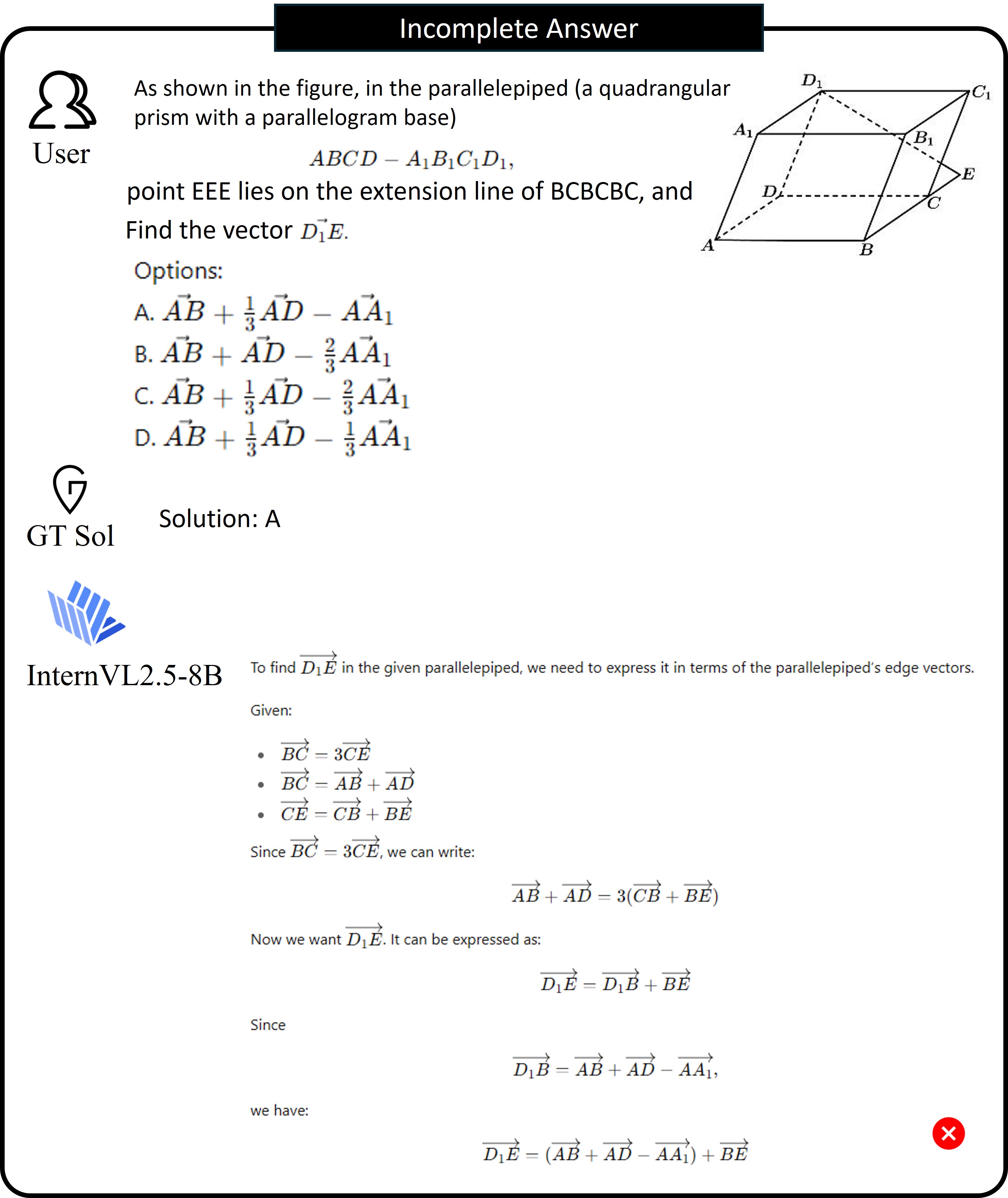}
    \caption{Error Analysis: Model failure case due to incomplete answer.}
    \label{fig:case_error_answer_incomplete}
\end{figure*}

\begin{figure*}
    \centering
    \centering
    \includegraphics[trim=0cm 0cm 0 0, width=0.9\textwidth]{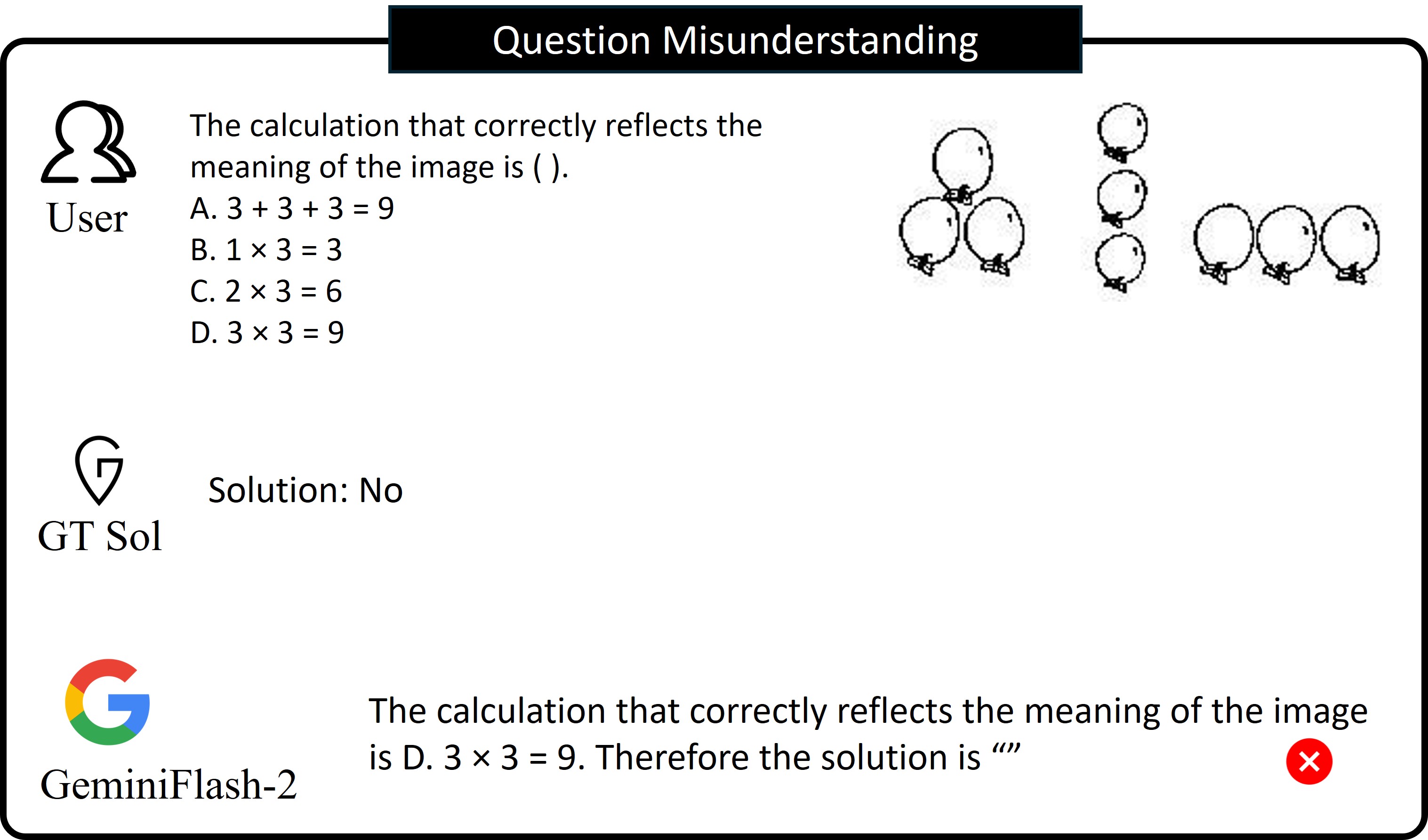}
    \caption{Error Analysis: Model failure case due to question misunderstanding.}
    \label{fig:case_error_question_miunderstanding}
\end{figure*}

\begin{figure*}
    \centering
    \centering
    \includegraphics[trim=0cm 0cm 0 0, width=0.9\textwidth]{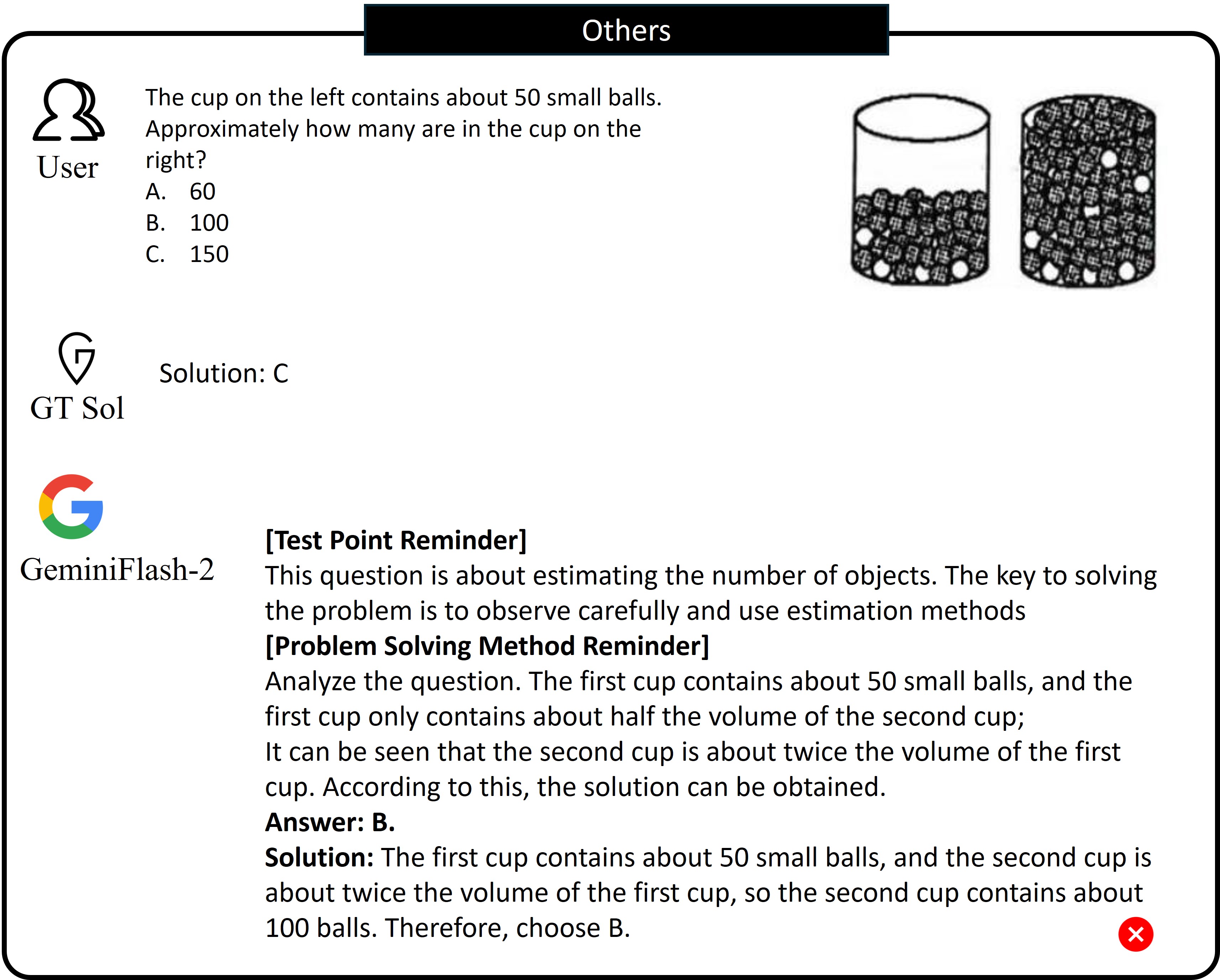}
    \caption{Error Analysis: Model failure case due to other issue. In this failure case, the model’s incorrect response stems from ambiguities or difficulties in judging (e.g., issues unrelated to reasoning, visual comprehension, or understanding of the question itself). For instance, the image provided in the question is unclear, making it difficult to determine whether the number of balls shown on the left represents 1/2 or 1/3 of the quantity shown on the right. Such uncertainty prevents the model from making a reliable comparison or calculation, leading to an incorrect or inconsistent answer.}
    \label{fig:other}
\end{figure*}









\bibliography{aaai2026}

\begin{thebibliography}{39}
\providecommand{\natexlab}[1]{#1}

\bibitem[{Alayrac et~al.(2022)Alayrac, Donahue, Luc, Miech, Barr, Hasson, Lenc, Mensch, Millican, Reynolds et~al.}]{alayrac2022flamingo}
Alayrac, J.-B.; Donahue, J.; Luc, P.; Miech, A.; Barr, I.; Hasson, Y.; Lenc, K.; Mensch, A.; Millican, K.; Reynolds, M.; et~al. 2022.
\newblock Flamingo: a visual language model for few-shot learning.
\newblock \emph{Advances in neural information processing systems}, 35: 23716--23736.

\bibitem[{Anthropic(2023)}]{claud3}
Anthropic. 2023.
\newblock The Claude 3 Model Family: Opus, Sonnet, Haiku.

\bibitem[{Arora, Singh et~al.(2023)}]{arora2023have}
Arora, D.; Singh, H.; et~al. 2023.
\newblock Have LLMs Advanced Enough? A Challenging Problem Solving Benchmark For Large Language Models.
\newblock In \emph{Proceedings of the 2023 Conference on Empirical Methods in Natural Language Processing}, 7527--7543.

\bibitem[{Bai et~al.(2025)Bai, Chen, Liu, Wang, Ge, Song, Dang, Wang, Wang, Tang et~al.}]{bai2025qwen2}
Bai, S.; Chen, K.; Liu, X.; Wang, J.; Ge, W.; Song, S.; Dang, K.; Wang, P.; Wang, S.; Tang, J.; et~al. 2025.
\newblock Qwen2. 5-vl technical report.
\newblock \emph{arXiv preprint arXiv:2502.13923}.

\bibitem[{{Black Forest Labs}(2024)}]{blackforestlabs_flux}
{Black Forest Labs}. 2024.
\newblock Flux.
\newblock \url{https://github.com/black-forest-labs/flux}.
\newblock Accessed: 2024-11-05.

\bibitem[{Chen et~al.(2024{\natexlab{a}})Chen, Li, Dong, Zhang, Zang, Chen, Duan, Wang, Qiao, Lin et~al.}]{chenwe}
Chen, L.; Li, J.; Dong, X.; Zhang, P.; Zang, Y.; Chen, Z.; Duan, H.; Wang, J.; Qiao, Y.; Lin, D.; et~al. 2024{\natexlab{a}}.
\newblock Are We on the Right Way for Evaluating Large Vision-Language Models?
\newblock In \emph{The Thirty-eighth Annual Conference on Neural Information Processing Systems}.

\bibitem[{Chen et~al.(2024{\natexlab{b}})Chen, Wang, Cao, Liu, Gao, Cui, Zhu, Ye, Tian, Liu et~al.}]{chen2024expandinginternvl2.5}
Chen, Z.; Wang, W.; Cao, Y.; Liu, Y.; Gao, Z.; Cui, E.; Zhu, J.; Ye, S.; Tian, H.; Liu, Z.; et~al. 2024{\natexlab{b}}.
\newblock Expanding performance boundaries of open-source multimodal models with model, data, and test-time scaling.
\newblock \emph{arXiv preprint arXiv:2412.05271}.

\bibitem[{Cobbe et~al.(2021)Cobbe, Kosaraju, Bavarian, Chen, Jun, Kaiser, Plappert, Tworek, Hilton, Nakano et~al.}]{cobbe2021training}
Cobbe, K.; Kosaraju, V.; Bavarian, M.; Chen, M.; Jun, H.; Kaiser, L.; Plappert, M.; Tworek, J.; Hilton, J.; Nakano, R.; et~al. 2021.
\newblock Training verifiers to solve math word problems.
\newblock \emph{arXiv preprint arXiv:2110.14168}.

\bibitem[{Das et~al.(2024)Das, Hristov, Li, Dimitrov, Koychev, and Nakov}]{das2024exams}
Das, R.~J.; Hristov, S.~E.; Li, H.; Dimitrov, D.~I.; Koychev, I.; and Nakov, P. 2024.
\newblock EXAMS-V: A Multi-Discipline Multilingual Multimodal Exam Benchmark for Evaluating Vision Language Models.
\newblock \emph{arXiv preprint arXiv:2403.10378}.

\bibitem[{Ding et~al.(2024)Ding, Deng, Choo, Wu, Agrawal, Schwarzschild, Zhou, Goldstein, Langford, Anandkumar et~al.}]{ding2024easy2hard}
Ding, M.; Deng, C.; Choo, J.; Wu, Z.; Agrawal, A.; Schwarzschild, A.; Zhou, T.; Goldstein, T.; Langford, J.; Anandkumar, A.; et~al. 2024.
\newblock Easy2Hard-Bench: Standardized Difficulty Labels for Profiling LLM Performance and Generalization.
\newblock \emph{Advances in Neural Information Processing Systems}, 37: 44323--44365.

\bibitem[{Hao et~al.(2025)Hao, Gu, Wang, Li, Yang, Wang, and Cheng}]{hao2025can}
Hao, Y.; Gu, J.; Wang, H.~W.; Li, L.; Yang, Z.; Wang, L.; and Cheng, Y. 2025.
\newblock Can MLLMs Reason in Multimodality? EMMA: An Enhanced MultiModal ReAsoning Benchmark.
\newblock In \emph{Forty-second International Conference on Machine Learning}.

\bibitem[{He et~al.(2024)He, Luo, Bai, Hu, Thai, Shen, Hu, Han, Huang, Zhang et~al.}]{he2024olympiadbench}
He, C.; Luo, R.; Bai, Y.; Hu, S.; Thai, Z.; Shen, J.; Hu, J.; Han, X.; Huang, Y.; Zhang, Y.; et~al. 2024.
\newblock OlympiadBench: A Challenging Benchmark for Promoting AGI with Olympiad-Level Bilingual Multimodal Scientific Problems.
\newblock In \emph{Proceedings of the 62nd Annual Meeting of the Association for Computational Linguistics (Volume 1: Long Papers)}, 3828--3850.

\bibitem[{Hendrycks et~al.(2021{\natexlab{a}})Hendrycks, Burns, Basart, Zou, Mazeika, Song, and Steinhardt}]{hendryckstest2021}
Hendrycks, D.; Burns, C.; Basart, S.; Zou, A.; Mazeika, M.; Song, D.; and Steinhardt, J. 2021{\natexlab{a}}.
\newblock Measuring Massive Multitask Language Understanding.
\newblock \emph{Proceedings of the International Conference on Learning Representations (ICLR)}.

\bibitem[{Hendrycks et~al.(2021{\natexlab{b}})Hendrycks, Burns, Kadavath, Arora, Basart, Tang, Song, and Steinhardt}]{hendrycks2measuring}
Hendrycks, D.; Burns, C.; Kadavath, S.; Arora, A.; Basart, S.; Tang, E.; Song, D.; and Steinhardt, J. 2021{\natexlab{b}}.
\newblock Measuring Mathematical Problem Solving With the MATH Dataset.
\newblock In \emph{Thirty-fifth Conference on Neural Information Processing Systems Datasets and Benchmarks Track (Round 2)}.

\bibitem[{Hu et~al.(2025)Hu, Gu, Dou, Fayyaz, Lu, Chang, and Peng}]{humrag}
Hu, W.; Gu, J.-C.; Dou, Z.-Y.; Fayyaz, M.; Lu, P.; Chang, K.-W.; and Peng, N. 2025.
\newblock MRAG-Bench: Vision-Centric Evaluation for Retrieval-Augmented Multimodal Models.
\newblock In \emph{The Thirteenth International Conference on Learning Representations}.

\bibitem[{Huang et~al.(2024)Huang, Wang, Xia, Li, Zou, Xu, Fan, Ye, Chern, Ye et~al.}]{huang2024olympicarena}
Huang, Z.; Wang, Z.; Xia, S.; Li, X.; Zou, H.; Xu, R.; Fan, R.-Z.; Ye, L.; Chern, E.; Ye, Y.; et~al. 2024.
\newblock Olympicarena: Benchmarking multi-discipline cognitive reasoning for superintelligent ai.
\newblock \emph{Advances in Neural Information Processing Systems}, 37: 19209--19253.

\bibitem[{Li et~al.(2025)Li, Zhu, Zhang, Lin, Zhou, and Xie}]{li2025k12vista}
Li, C.; Zhu, C.; Zhang, T.; Lin, M.; Zhou, Z.; and Xie, J. 2025.
\newblock K12Vista: Exploring the Boundaries of MLLMs in K-12 Education.
\newblock \emph{arXiv preprint arXiv:2506.01676}.

\bibitem[{Li et~al.(2022)Li, Li, Xiong, and Hoi}]{li2022blip}
Li, J.; Li, D.; Xiong, C.; and Hoi, S. 2022.
\newblock Blip: Bootstrapping language-image pre-training for unified vision-language understanding and generation.
\newblock In \emph{International conference on machine learning}, 12888--12900. PMLR.

\bibitem[{Liu et~al.(2024{\natexlab{a}})Liu, Li, Wu, and Lee}]{liu2023llava}
Liu, H.; Li, C.; Wu, Q.; and Lee, Y.~J. 2024{\natexlab{a}}.
\newblock Visual instruction tuning.
\newblock \emph{Advances in neural information processing systems}, 36.

\bibitem[{Liu et~al.(2024{\natexlab{b}})Liu, Duan, Zhang, Li, Zhang, Zhao, Yuan, Wang, He, Liu et~al.}]{liu2024mmbench}
Liu, Y.; Duan, H.; Zhang, Y.; Li, B.; Zhang, S.; Zhao, W.; Yuan, Y.; Wang, J.; He, C.; Liu, Z.; et~al. 2024{\natexlab{b}}.
\newblock Mmbench: Is your multi-modal model an all-around player?
\newblock In \emph{European conference on computer vision}, 216--233. Springer.

\bibitem[{Lohman and Lakin(2011)}]{lohman2011intelligence}
Lohman, D.~F.; and Lakin, J.~M. 2011.
\newblock Intelligence and reasoning.
\newblock \emph{The Cambridge handbook of intelligence}, 419--441.

\bibitem[{Lu et~al.(2022)Lu, Mishra, Xia, Qiu, Chang, Zhu, Tafjord, Clark, and Kalyan}]{lu2022learn}
Lu, P.; Mishra, S.; Xia, T.; Qiu, L.; Chang, K.-W.; Zhu, S.-C.; Tafjord, O.; Clark, P.; and Kalyan, A. 2022.
\newblock Learn to explain: Multimodal reasoning via thought chains for science question answering.
\newblock \emph{Advances in Neural Information Processing Systems}, 35: 2507--2521.

\bibitem[{OpenAI(2024{\natexlab{a}})}]{openai2024gpt4o}
OpenAI. 2024{\natexlab{a}}.
\newblock GPT-4o: A Multimodal Language Model.
\newblock Accessed: 2025-03-08.

\bibitem[{OpenAI(2024{\natexlab{b}})}]{openai2024gpto1mini}
OpenAI. 2024{\natexlab{b}}.
\newblock GPT-o1-mini: A Multimodal Language Model.
\newblock Accessed: 2025-03-08.

\bibitem[{Radford et~al.(2021)Radford, Kim, Hallacy, Ramesh, Goh, Agarwal, Sastry, Askell, Mishkin, Clark et~al.}]{radford2021learning}
Radford, A.; Kim, J.~W.; Hallacy, C.; Ramesh, A.; Goh, G.; Agarwal, S.; Sastry, G.; Askell, A.; Mishkin, P.; Clark, J.; et~al. 2021.
\newblock Learning transferable visual models from natural language supervision.
\newblock In \emph{International conference on machine learning}, 8748--8763. PmLR.

\bibitem[{Rein et~al.(2025)Rein, Hou, Stickland, Petty, Pang, Dirani, Michael, and Bowman}]{rein2024gpqa}
Rein, D.; Hou, B.~L.; Stickland, A.~C.; Petty, J.; Pang, R.~Y.; Dirani, J.; Michael, J.; and Bowman, S.~R. 2025.
\newblock Gpqa: A graduate-level google-proof q\&a benchmark.
\newblock In \emph{First Conference on Language Modeling}.

\bibitem[{Sternberg(1982)}]{sternberg1982reasoning}
Sternberg, R.~J. 1982.
\newblock Reasoning, problem solving, and intelligence.
\newblock \emph{Handbook of human intelligence}, 225--307.

\bibitem[{Sun et~al.(2024{\natexlab{a}})Sun, Han, Zhao, Ma, Shen, Chen, Chen, and Yu}]{sun2024scieval}
Sun, L.; Han, Y.; Zhao, Z.; Ma, D.; Shen, Z.; Chen, B.; Chen, L.; and Yu, K. 2024{\natexlab{a}}.
\newblock Scieval: A multi-level large language model evaluation benchmark for scientific research.
\newblock In \emph{Proceedings of the AAAI Conference on Artificial Intelligence}, volume~38, 19053--19061.

\bibitem[{Sun et~al.(2024{\natexlab{b}})Sun, Wu, Zhu, Zheng, Chen, Zhang, Zhang, Wan, Lan, Zheng et~al.}]{sun2024pathmmu}
Sun, Y.; Wu, H.; Zhu, C.; Zheng, S.; Chen, Q.; Zhang, K.; Zhang, Y.; Wan, D.; Lan, X.; Zheng, M.; et~al. 2024{\natexlab{b}}.
\newblock Pathmmu: A massive multimodal expert-level benchmark for understanding and reasoning in pathology.
\newblock In \emph{European Conference on Computer Vision}, 56--73. Springer.

\bibitem[{Team et~al.(2023)Team, Anil, Borgeaud, Alayrac, Yu, Soricut, Schalkwyk, Dai, Hauth, Millican et~al.}]{team2023gemini}
Team, G.; Anil, R.; Borgeaud, S.; Alayrac, J.-B.; Yu, J.; Soricut, R.; Schalkwyk, J.; Dai, A.~M.; Hauth, A.; Millican, K.; et~al. 2023.
\newblock Gemini: a family of highly capable multimodal models.
\newblock \emph{arXiv preprint arXiv:2312.11805}.

\bibitem[{Team(2024)}]{qvq-72b-preview}
Team, Q. 2024.
\newblock QVQ: To See the World with Wisdom.

\bibitem[{Wang et~al.(2024{\natexlab{a}})Wang, Chen, Wang, Cao, Liu, Gao, Zhu, Zhu, Lu, Qiao, and Dai}]{wang2024mpo}
Wang, W.; Chen, Z.; Wang, W.; Cao, Y.; Liu, Y.; Gao, Z.; Zhu, J.; Zhu, X.; Lu, L.; Qiao, Y.; and Dai, J. 2024{\natexlab{a}}.
\newblock Enhancing the Reasoning Ability of Multimodal Large Language Models via Mixed Preference Optimization.
\newblock \emph{arXiv preprint arXiv:2411.10442}.

\bibitem[{Wang et~al.(2024{\natexlab{b}})Wang, Hu, Lu, Zhu, Zhang, Subramaniam, Loomba, Zhang, Sun, and Wang}]{wang2024scibench}
Wang, X.; Hu, Z.; Lu, P.; Zhu, Y.; Zhang, J.; Subramaniam, S.; Loomba, A.~R.; Zhang, S.; Sun, Y.; and Wang, W. 2024{\natexlab{b}}.
\newblock SciBench: Evaluating College-Level Scientific Problem-Solving Abilities of Large Language Models.
\newblock In \emph{International Conference on Machine Learning}, 50622--50649. PMLR.

\bibitem[{Xu et~al.(2024)Xu, Wang, Fan, and Liu}]{xu2024benchmarking}
Xu, R.; Wang, Z.; Fan, R.-Z.; and Liu, P. 2024.
\newblock Benchmarking benchmark leakage in large language models.
\newblock \emph{arXiv preprint arXiv:2404.18824}.

\bibitem[{Yang et~al.(2024)Yang, Zhang, Shao, Zhang, Bin, Wang, and Luo}]{yang2024dynamic}
Yang, Y.; Zhang, S.; Shao, W.; Zhang, K.; Bin, Y.; Wang, Y.; and Luo, P. 2024.
\newblock Dynamic Multimodal Evaluation with Flexible Complexity by Vision-Language Bootstrapping.
\newblock \emph{arXiv preprint arXiv:2410.08695}.

\bibitem[{Yue et~al.(2024)Yue, Ni, Zhang, Zheng, Liu, Zhang, Stevens, Jiang, Ren, Sun et~al.}]{yue2024mmmu}
Yue, X.; Ni, Y.; Zhang, K.; Zheng, T.; Liu, R.; Zhang, G.; Stevens, S.; Jiang, D.; Ren, W.; Sun, Y.; et~al. 2024.
\newblock {MMMU}: A massive multi-discipline multimodal understanding and reasoning benchmark for expert {AGI}.
\newblock In \emph{Proceedings of the IEEE/CVF Conference on Computer Vision and Pattern Recognition}, 9556--9567.

\bibitem[{Zhang et~al.(2024)Zhang, Jiang, Zhang, Lin, Guo, Qiu, Zhou, Lu, Chang, Qiao et~al.}]{zhang2024mathverse}
Zhang, R.; Jiang, D.; Zhang, Y.; Lin, H.; Guo, Z.; Qiu, P.; Zhou, A.; Lu, P.; Chang, K.-W.; Qiao, Y.; et~al. 2024.
\newblock Mathverse: Does your multi-modal llm truly see the diagrams in visual math problems?
\newblock In \emph{European Conference on Computer Vision}, 169--186. Springer.

\bibitem[{Zhong et~al.(2024)Zhong, Cui, Guo, Liang, Lu, Wang, Saied, Chen, and Duan}]{zhong2024agieval}
Zhong, W.; Cui, R.; Guo, Y.; Liang, Y.; Lu, S.; Wang, Y.; Saied, A.; Chen, W.; and Duan, N. 2024.
\newblock AGIEval: A Human-Centric Benchmark for Evaluating Foundation Models.
\newblock In \emph{NAACL-HLT (Findings)}.

\bibitem[{Zhu et~al.(2024)Zhu, Wang, Zhao, Xu, and Xie}]{zhu2024dynamic}
Zhu, K.; Wang, J.; Zhao, Q.; Xu, R.; and Xie, X. 2024.
\newblock Dynamic Evaluation of Large Language Models by Meta Probing Agents.
\newblock In \emph{Forty-first International Conference on Machine Learning}.

\end{thebibliography}

\end{document}